%% file: main.tex
\documentclass[lettersize,journal]{IEEEtran}
\usepackage{amsmath,amsfonts}
\usepackage{array}
\usepackage[caption=false,font=normalsize,labelfont=sf,textfont=sf]{subfig}
\usepackage{textcomp}
\usepackage{stfloats}
\usepackage{url}
\usepackage{verbatim}
\usepackage{graphicx}
\usepackage[backend=biber]{biblatex}
\usepackage{mathtools}
\DeclarePairedDelimiter\ceil{\lceil}{\rceil}

\usepackage[ruled,vlined]{algorithm2e}

\usepackage{tabu}                      
\usepackage{booktabs}                  

\usepackage{comment}
\usepackage{xspace}
\usepackage{tabularx}
\usepackage[subtle]{savetrees}  
\usepackage{lettrine}

\SetCommentSty{mycommfont}
\newcount\notenum
\newcommand{\fslnote}[3]        {\global\advance\notenum by 1\textsl{\bf\color{#2}[#1\the\notenum: #3]}}

\newcommand{\mario}[1]          {\fslnote{Mario}{DarkGreen}{#1}}

\newcommand{\ie}                {\emph{i.e.},\xspace}
\newcommand{\eg}                {\emph{e.g.},\xspace}
\newcommand{\etal}              {\emph{et~al.}\xspace}

\begin{document}

\title{Into the Void: Mapping the Unseen Gaps in High Dimensional Data}

\author{ Xinyu Zhang, Tyler Estro, 
  Geoff Kuenning,~\IEEEmembership{Life Member, IEEE}, \\
  Erez Zadok,~\IEEEmembership{Member, IEEE}, 
  Klaus Mueller,~\IEEEmembership{Fellow, IEEE}
\thanks{Xinyu Zhang, Tyler Estro, Erez Zadok and Klaus Mueller are with the Department of Computer Science, Stony Brook University, New York.
  	E-mail: \{zhang146\,$|$\, testro\,$|$\, ezk\,$|$\, mueller\}@cs.stonybrook.edu\,.}
\thanks{Geoff Kuenning is with the Department of Computer Science, Harvey Mudd College, Claremont, California.
  	E-mail: geoff@cs.hmc.edu\,.}}

\markboth{Journal of \LaTeX\ Class Files,~Vol.~14, No.~8, August~2021}%
{Shell \MakeLowercase{\textit{et al.}}: A Sample Article Using IEEEtran.cls for IEEE Journals}


\maketitle

\input{abstract}


\input{intro}
\input{related-work}

\begin{figure}[tbp]
  \centering
  \includegraphics[width=\columnwidth]{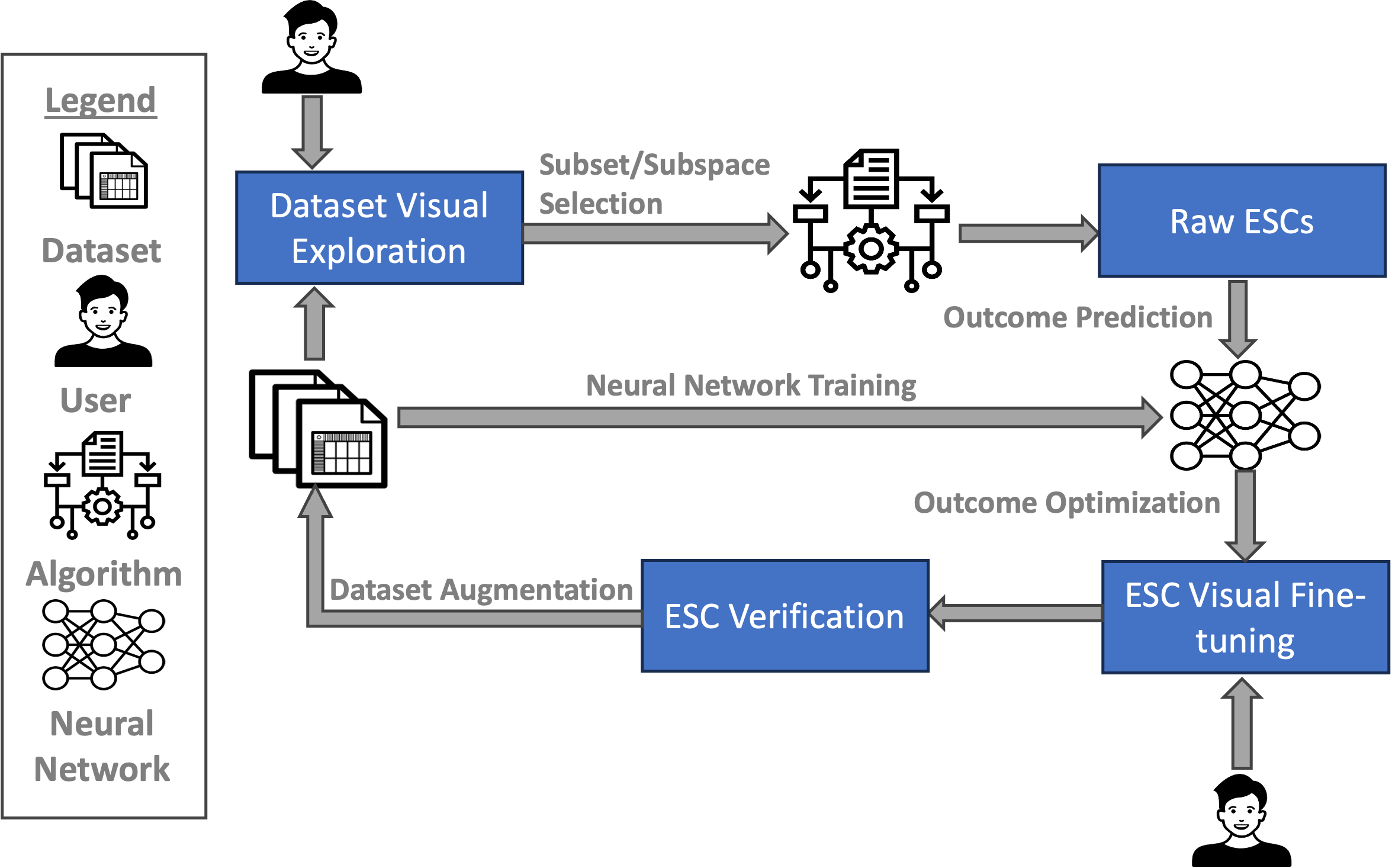}
    \caption{Our three-phase workflow.}
  \hfill%
  \vspace{-20 pt}
  \label{fig:workflow}
\end{figure}

\section{Overview}
Fig. \ref{fig:workflow} presents an overview of our methodology. It comprises three phases centered around training an assistive deep neural network (DNN) that becomes increasingly adept at recommending useful empty space configurations (ESCs) for the target (domain) application's possibly multiple objectives. The process begins with a fully untrained DNN, where the user (typically a domain expert) and our empty space search algorithm (ESA) collaborate to identify initial "raw" ESC candidates; this is depicted in the top branch of Fig. \ref{fig:workflow}. These candidates are then interactively refined by the user using visual tools and applied within the target application, as shown in the bottom branch of Fig. \ref{fig:workflow}. The application is run with these configurations, and their performance is measured. Once verified, these ESCs augment the dataset and are used to train the DNN (per the middle branch of Fig. \ref{fig:workflow}). As the DNN improves, the user's involvement gradually decreases. 

Ultimately, the pipeline operates autonomously: the ESA identifies ESC candidates, the DNN optimizes and approximates the target application's outcomes, and the dataset is continuously updated with new ESCs. In the following we describe the various components of this workflow. 

\section{Empty Space Search Algorithm}
\label{sec:es_algorithm}
The search for empty regions in a high-D data space is a main premise in our work. The challenge here is to describe this space effectively, without enumerating all of the points that reside within each continuous (empty) region. 
A key requirement is that interior points within an empty space should be far from known points. To locate the emptiest region within a group of data points, one could use Delaunay Triangulation, where the circumscribed center represents the emptiest point. However, as mentioned, computational-geometry methods face the curse of dimensionality.

We instead propose an agent-based approach. Here, the agent is repelled from known data points if it comes too close and is attracted back if it strays too far. This premise is the aim of a physics-inspired function called the \textit{Lennard-Jones Potential}, which is the principle guiding our heuristic search algorithm. A particular advantage of this scheme is that it is easily parallelizable. Agents can be deployed throughout the high-D data space and operate autonomously to identify empty space points. Next, we describe this physics-inspired search method in detail.

\subsection{Physics Background: The Lennard-Jones Potential}

\begin{figure}[tb]
  \centering 
  \includegraphics[width=0.55\columnwidth]{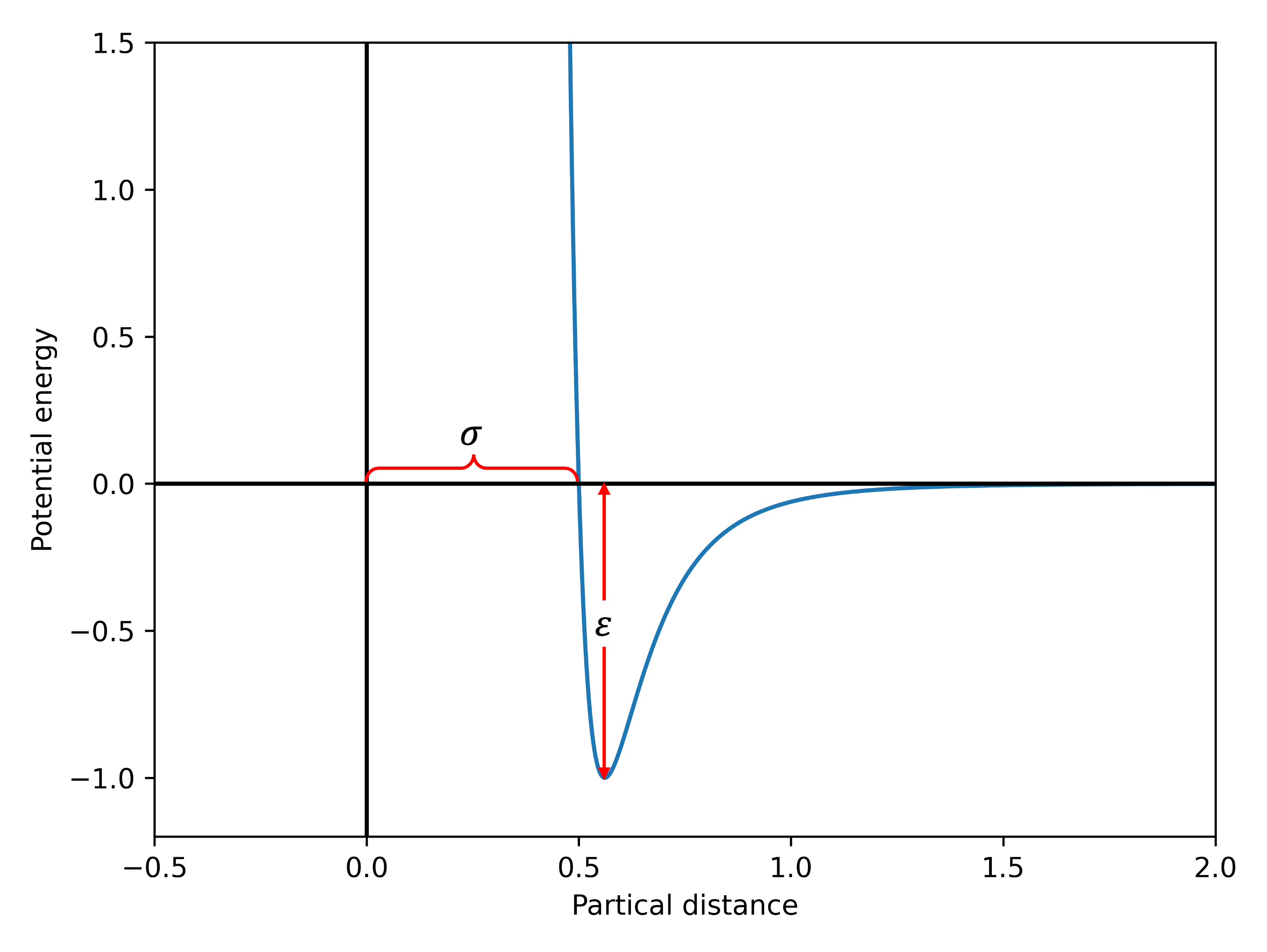}
  \vspace{-5 pt}
  \caption{
    An example of Lennard-Jones Potential. The $x$ axis is the
    distance between particles ($r$) and the y axis is the outcome
    $V(r)$. The potential is positive when the particle distance is
    less than $\sigma$, indicating a repulsive force, and negative for
    larger distances, where there is an attractive force. The minimum potential (strongest attraction) is
    $-\epsilon$.
  }
  \label{fig:LJ}
  \vspace{-10 pt}
\end{figure}

There are multiple causes of intermolecular interactions, including Van der
Waals forces, hydrogen bonds, and electrostatic interactions. Some of
the forces are repulsive and others are attractive, so that a dynamic
equilibrium of molecules is maintained. The Lennard-Jones (L-J) Potential
\cite{lennard1924determination} (see Fig. \ref{fig:LJ} for an example) is a popular
model of intermolecular interactions and is described as:

\begin{equation}
    \label{eq:lj}
    V(r) = 4\epsilon \left[(\frac{\sigma}{r})^{12} - ({\frac{\sigma}{r})^6}\right]
\end{equation}
where $r$ and $V(r)$ are the distance and potential of a
pair of particles, and $\epsilon$ represents the depth of the potential
well, correlating to the strength of the interaction between two
particles. $\sigma$ represents the effective diameter of the
particles, that is, the distance at which the total potential energy
between two particles becomes zero.
At distances less than $\sigma$, the repulsive force dominates, causing the potential energy to increase sharply. At distances greater than $\sigma$, the attractive force is stronger, pulling the particles together, but this force diminishes with $r$. The nature of intermolecular interactions keeps a point
dynamically steady in a local region. Our search algorithm uses the L-J Potential function. 

\subsection{Our Lennard-Jones Potential Based Search Algorithm}

We assume that there is a possibly small initial dataset with verified
configurations, and we populate the high-D search space with these
known configurations.
Then we place an agent in this space to search for Empty Space
Configurations (ESC)---the raw ESCs in Fig. \ref{fig:workflow}. The agent
starts from a randomly sampled initial position and uses the L-J
Potential function to move to a location where the potential becomes
zero. This trajectory is sampled along the way to form a set of raw
ESCs.

The vector
intermolecular force $F(r)$ driving the agent is:
\begin{equation}
    \label{eq:flj}
    F(r) = \frac{dV(r)}{dr} = 24\frac{\epsilon}{\sigma}\left[2\left(\frac{\sigma}{r}\right)^{13} - \left(\frac{\sigma}{r}\right)^7\right]
\end{equation}

\noindent 
To adhere to physical reality, we use the resultant force
$\Sigma \Vec{F}$ to determine the direction of motion $\Vec{d}$, rather than
simply using $V(r)$. However, we will stop the agent if $||\Sigma
\Vec{F}||$ falls below a threshold as this indicates it has moved
beyond the data manifold. $\Sigma \Vec{F}$ is the sum of forces due to
the agent's $k$ nearest neighbors in the dataset:
\begin{equation}
    \label{eq: sumf}
    \begin{aligned}
        \Sigma \Vec{F} &= \sum_{i}^{k}\Vec{u_i}F(r_i) \qquad
        \Vec{d} &= \frac{\Sigma \Vec{F}}{||\Sigma \Vec{F}||}
    \end{aligned}
\end{equation}
where $k$ is the number of neighbors and $\Vec{u_i}$ is the unit
vector from the agent to the $i$th neighbor.

\begin{algorithm}
\SetAlgoLined
 Set the number of neighbors $k$, the particle effective diameter $\sigma$, the number of search steps $n$, the
 step size $\alpha$, the discount factor $\gamma$, the vanishing threshold $\delta$ and the rollout interval $j$\;

 Initialize a trajectory $\tau = []$ and an agent $\pi = \mathbf{c}$
 where $\mathbf{c}$ is a random coordinate; set
 cumulative magnitude $L=0$, momentum $\Vec{m} = \Vec{0}$\;
 Specify constraints on agent:\\
 \hspace*{0.5em} $f_1(\pi) <= 0$; $f_2(\pi) <= 0$; \dots, $f_p(\pi) <= 0$\;
 \For {i \dots $n$} {
    \If{i \% $j$ == 0} {
        Add current coordinate of $\pi$ to $\tau$\; 
    }
    Get $k$ nearest neighbors of the agent from the dataset\;
    Use Eq. \ref{eq: sumf} to calculate $||\Sigma \Vec{F}||$ and $\Vec{d}$\;
    \If {$||\Sigma \Vec{F}|| < \delta$} {
        return $\tau$\;
    } 

        $
            \Vec{d} = (\Vec{d} * ||\Sigma \Vec{F}|| + \Vec{m} * L) / (L + ||\Sigma \Vec{F}||)
        $\;
        $\pi = \pi + \Vec{d} * \alpha$\;
        $L = \gamma * L + ||\Sigma \Vec{F}||$\;
        $\Vec{m} = \gamma * \Vec{m} + \Vec{d}$\;
        $\Vec{m} = \Vec{m} / ||\Vec{m}||$\;
    
    
    \If{$\pi$ violates any constraint $f$} {
        return $\tau$\;
    }
 }
 return $\tau$\;
 \caption{Empty-Space Search for a Single Agent}
 \label{alg:search}
\end{algorithm}

Alg. \ref{alg:search} illustrates the details of our empty-space search. It returns the
trajectory $\tau$ of an agent, where each element is a raw ESC. Users
can set two accuracy levels for the DNN, defined by the DNN's
acceptable prediction error, measured in \%. When the DNN performs
worse than the lower accuracy level, which occurs in the initial
learning phase, it cannot reliably estimate the value of an ESC. In
this case we set 
$\gamma=0$,
causing the agent $\pi$ to simply move in the direction of $\Vec{d}$;
when converged it returns the end of $\tau$ as the ESC. This strategy
encourages the algorithm to identify as many gaps as possible,
fostering the development of a robust neural network and accelerating
training. However, it may also lead to convergence on local minima.

The time complexity of ESA is $O(dknp + np\log N)$ where $d$ is the dimensionality of the dataset, $k$ is the number of nearest neighbors, $n$ is the number of search steps, $p$ is the number of agents, and $N$ is the size of the dataset. $O(dknp)$ represents the time complexity to determine the next step, while $O(np\log N)$ represents the time complexity to query $k$ nearest neighbors; the space complexity of ESA is $O(dp)$. In contrast, the time and space complexity of Delaunay Triangulation is $O(N^{\ceil{{d/2}}})$, as mentioned before. ESA is considerably more efficient in both time and space.

Once the DNN has developed further (phase 2 in the workflow of
Fig. \ref{fig:workflow}) and meets the lower accuracy level,
we begin incorporating a momentum $\Vec{m}$ with factor $\gamma > 0$ to move the agent along. This mechanism considers historical directions alongside those calculated by Eq. \ref{eq: sumf}, resulting in a smoother update for the direction of $\pi$. We constrain $\gamma < 1$ to prevent a long-term
  effect; thus, the effect of historical forces gets lower and lower
  with each search step. 
The momentum mechanism encourages the agent $\pi$ to explore the space in
a broader range before converging to the empty region, leading it to discover global minima. Then, upon terminating it returns $\tau$ as a list of ESCs. 
Eventually the DNN is accurate enough to achieve the higher accuracy level. We then 
improve $\tau$ by DNN-enabled gradient ascent and pick the best result. 

\begin{figure}[t]
    \centering
    \subfloat[]{\includegraphics[width=0.45\columnwidth]{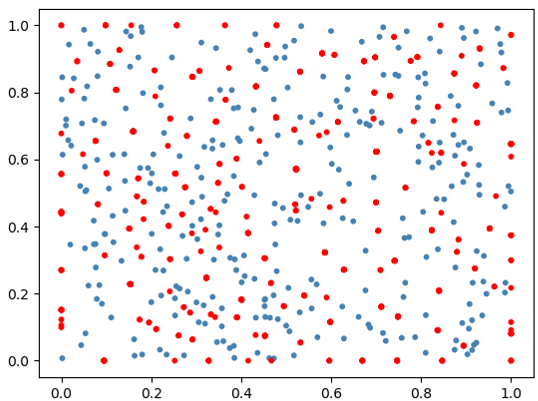}}
    \hfil
    \subfloat[]{\includegraphics[width=0.45\columnwidth]{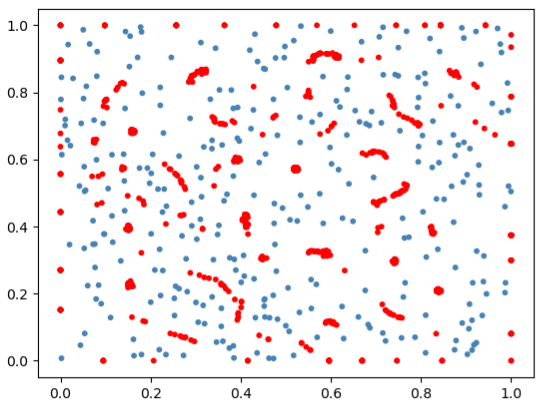}}
    \caption{Given 300 random samples (blue) and 600 random agents (red) in 2D space, we show the results of ESA (a) without and (b) with momentum.}
    \label{fig:ESA-points}
    \vspace{-10pt}
\end{figure}

Fig. \ref{fig:ESA-points} shows our algorithm's outcomes for
a 2D dataset. Agents tend to
converge to a small region without momentum (Fig. \ref{fig:ESA-points}(a)), but to a larger region with momentum (Fig. \ref{fig:ESA-points}(b)). Larger regions are explored more thoroughly in the latter case while smaller gaps are left alone.

We note that the ESA works in local space with simple calculations and scales well to higher dimensions and larger datasets (see also Sec. \ref{sec:case}). As mentioned, since each agent operates independently, it is feasible to expedite the empty space search using a GPU and deploy a large batch of agents concurrently. DT does not offer this kind of parallelism.

\section{GapMiner}
\label{sec:ui_design}

Our visual analytics system, GapMiner, combines data space exploration and HITL to aid users in identifying promising ESCs during the first two phases of the workflow when the DNN is not yet fully trained.
\subsection{Design Goals}

Following the design model devised by
Munzner\cite{munzner2009nested}, we studied popular datasets on Kaggle,
reviewed recent research in our example domain (computer systems
\cite{waldspurger2015efficient}), and interviewed several domain
experts.
This study yielded the essential features an effective visual analytics system for empty-space search should have:

\begin{itemize}
    \item \textbf{F1: Subset/Subspace selection.}  Users often focus
        on specific data segments; e.g., a system designer with a
        limited budget might want to explore moderate options
        first. Sometimes these subsets are so small that their key
        features are hidden within the entire dataset. Hence,
        explorations within data subspaces must be supported.

    \item \textbf{F2: Data distribution visualization and cluster highlighting.}     
    For data with categorical outcomes, each label typically forms a
    distinct cluster. For continuous outcomes, high and low values
    usually don't overlap. Highlighting clusters and visualizing
    distributions will help users identify valuable empty spaces for
    exploration.

    \item \textbf{F3: HITL ESC fine-tuning.} While the ESA identifies numerous ESCs, they are not fully refined configurations, but only starting points for exploration. Users should be able to define empty space search regions and refine raw ESCs based on their expertise.
    
    \item \textbf{F4: ESC neighbor visualization.} In order to select
        promising ESCs, users need to understand the distribution of
        their neighbors, \ie the shape of the empty space
        (hypersphere, paraboloid, hyperbolas, etc.),
to aid informed decision making.
    
    \item \textbf{F5: Outcome evaluation.} Users must be able to evaluate the performance of an ESC which sometimes is gauged by more than a single outcome variable. Additionally, the cost of obtaining certain configurations can also be important.
\end{itemize}

\begin{figure*}
    \centering
  \includegraphics[width=0.9\linewidth]{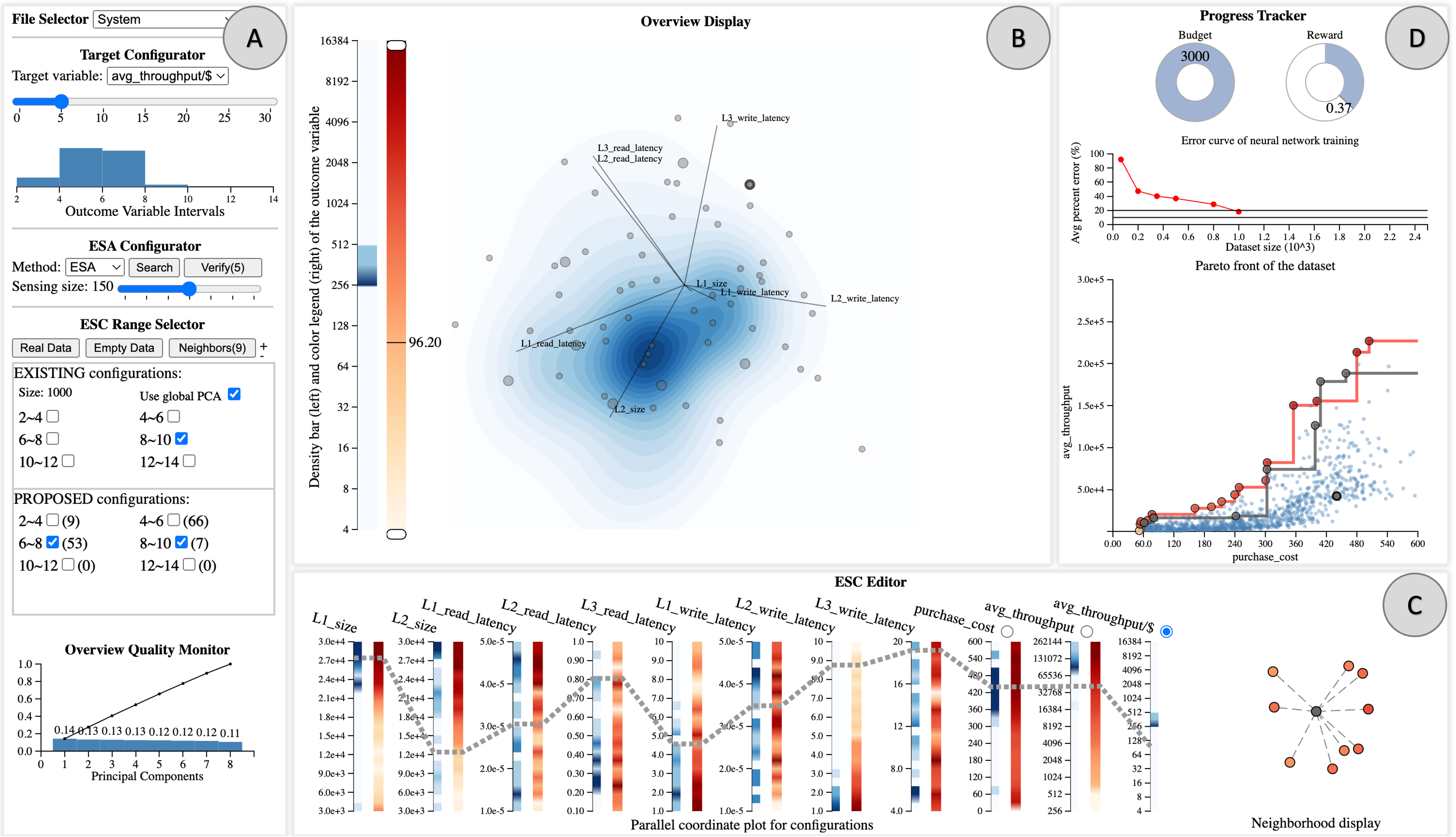}
  \caption{GapMiner visual interface where a selected ESC is reflected
    in all displays. (A) Control Panel. From top to bottom: (a) File
    Selector to load a dataset of initial verified configurations with
    values for all parameter variables. (b) Target Variable
    Configurator with an interface for breaking its value range into
    discrete intervals. (c) Empty Space Algorithm (ESA) Configurator
    to select the ESA and a slider to set the ESC batch size. (d)
    Empty Space Configuration (ESC) Range Selector to control which
    target variable intervals are used for display and ESC
    proposals. (e) Overview Quality Monitor screeplot that shows the
    amount of data variance captured by the Overview (PCA)
    Display. (B) Overview (PCA) Display with data distribution
    contours, raw or modified ESCs rendered as points,  and color
    legends. (C) Empty Space Configuration (ESC) Editor.  From left to
    right: (a) Parallel Coordinate Plot where users can configure ESCs
    starting from a raw ESC or an existing configuration. (b) Neighbor
    plot of the selected ESC providing a local view of the
    distribution of its nearest existing configurations. (D) Progress
    Tracker. From top to bottom: (a) Budget/Reward Display that
    captures the aggregated evaluation cost and merit of the ESC
    exploration so far. (b) Training Status Display of the assistive
    DNN. (c) Pareto Frontier Plot that shows the Pareto frontiers of
    existing
configurations (red) and ESCs (gray) with respect to two user-chosen merit (target)  variables. }
  
  \label{fig:teaser}
  \vspace{-10pt}
\end{figure*}

\subsection{Terms, Color Semantics, and Representations}
\label{sec:termscolors}
 Fig. \ref{fig:teaser} shows the GapMiner interface. The dataset displayed consists of a set of multi-tier cache configurations used to replay a real-world block I/O trace, each with three devices: $L1$, $L2$, and $L3$. Each cache device has three variables: $size$ and \textit{average read/write latencies} (the size of the backend storage device $L3$ is fixed for all configurations and so it is not included). See Sec. \ref{sec:application} for more detail. The target variables are \textit{average throughput} and \textit{total purchase cost}, used to evaluate the performance and cost of each configuration, respectively. 

The term \textit{existing configurations} used in Fig. \ref{fig:teaser}
(A) refers to configurations in the dataset that have been verified, whereas
\textit{proposed configurations} are configurations proposed by the various available ESC search algorithms but
which have not been verified yet. GapMiner distinguishes these two types with
different colors. All gray points and lines in the interface
represent proposed configurations, while everything related to existing
configurations is colored blue, including density contours in
Fig. \ref{fig:teaser} (B), histogram bars in Fig. \ref{fig:teaser} (A) and
Fig. \ref{fig:teaser} (C), and scatter points in Fig. \ref{fig:teaser}
(B). Additionally, we apply an orange-to-red colormap in
Fig. \ref{fig:teaser} (D) to encode target variables in the Pareto
front of existing configurations, and the same colormap in
Fig. \ref{fig:teaser} (C) to encode target variables of existing
neighbors around the proposed one.

Zhang et al.~\cite{zhang2022graphical} illustrated the benefits of
contour maps in exemplar identification while Li et
al.~\cite{li2023scatteruq} combined density contours and scattered points for
outlier examination. In the PCA map in Fig. \ref{fig:teaser} (B), we
follow their design pattern, employing density contours to abstract
existing configurations and scattered points to denote proposed
configurations. In the PCP, we visualize proposed configurations as dashed
lines and existing configurations as solid lines shown in
Fig. \ref{fig:teaser} (C).

Next we describe all of GapMiner's components in detail, referring to the Interface in Fig. \ref{fig:teaser} and the Design Goals they satisfy.

\subsection{Control Panel}

After loading the dataset containing the initial verified configurations with values for all variables, the user will head to the Target Configurator (A) to select a target variable subject to optimization. The target variable can either be a native outcome variable such as \textit{performance} or \textit{cost}, or it can be a ratio like \textit{performance/cost}. The latter provides quick insights into efficiency, while the former two can be optimized within our multi-objective optimization to account for the user's preferences. The user can now utilize the slider below to evenly split the range (F1) of the selected target variable. This action divides the dataset according to the selected value interval and the histogram below visualizes the distribution (F2). The user can now choose among one of three empty space search methods from the dropdown menu: the physics-based ESA described above, random sampling, Pareto improvement, and a baseline (see Sec. \ref{sec:pipeline}).  

The ESC Range Selector enables users to select one or more subsets of
the data, as specified in the Target Configurator (see above), using
the checkboxes in the ``existing''/``proposed'' group (F1). The
``Size'' in the verified ``existing'' group indicates the number of
these configurations.

 \subsection{Overview Display}

The PCA-based Overview Display (B) linearly transforms the high-dimensional data space into a variance-maximizing 2D projection. It is an intuitive way to visualize data distributions and identify clusters (F2), We chose the linear dimension reduction provided by PCA since evaluating results from nonlinear methods like tSNE is challenging and connecting their embedding space to the original space is difficult. 

Once a checkbox is selected in the ESC Range Selector, the associated
subset of the data will be displayed in the Overview Display either as
density contours (``existing'') or as scattered points (``proposed''
or selected via other means). The scented color legend bars on the
left map values to color:
the left one represents the target variable density of the chosen subset,  while the right one shows the value range of the entire dataset.
The ``Use global PCA'' checkbox in the ESC Range Selector (A)
determines whether the PCA layout is applied to the chosen data subset
(F2) or the entire dataset; the PCA map will update
accordingly. Applying the PCA layout to the currently selected data
subset will reduce the layout loss and make the display more accurate
and focused. This improvement is quantified in the Overview Monitor
scree plot.


The loading vectors in the PCA display represent the projection of Cartesian axes from the original space, forming a biplot. Due to the linear nature of PCA, any value change along one axis in the original high-D space results in a proportional update along the corresponding loading vector in the PCA map. Additionally, the loading vector projection is translation-invariant; any point can be chosen as an anchor to project the Cartesian basis. We derive this mechanism
in closed form in 
App. \ref{appendix:pca_insights}.

\subsection{Empty Space Configuration (ESC) Editor}
This interface panel has two displays: a PCP (left) where an ESC can be configured and refined and a neighborhood plot (right) that shows the topology of the ESC's immediate point neighborhood. 

\textbf{PCP display.} The PCP allows users to fine-tune a proposed ESC or propose one on their own. It is the most effective display for this task as it provides simultaneous access to all variables and allows for easy adjustments through simple mouse interactions. For additional insight we provide two scented bars along each PCP axis (F1, F2). 

The (blue) density bar next to each axis shows that variable's value
distribution in the currently chosen data subset, with darker blue
indicating higher density in that range. This visual information
complements the density contours in the Overview Display (B). The
correlation bar next to the density bar shows the relationship between
the variable and the selected target variable, calculated from all
existing configurations. Users can select the target variable for this
calculation by clicking the radio button along its corresponding axis
(for example in (C) there are three such variables---two native
variables and one
ratio variable). For a linear relationship, the orange-red bar along
each variable axis indicates whether the correlation with the selected
target is positive or negative. For a nonlinear relationship, it
reveals which value interval corresponds to a higher target value
(darker color) and which to a lower target value (lighter orange).


Lastly, when users modify an ESC in the PCP, its 2D position in the Overview Display is updated proportionally along the direction of the corresponding loading vector. This is particularly useful when there is an empty space in the Overview Display, providing users with important feedback that they are on the right track.


\textbf{Neighborhood Plot.} This display, located to the right of the
PCP, helps users understand the shape of an empty space as well as the
neighbor distribution of the associated ESC (F4). To create this
visualization we devised a dedicated dimension-reduction method we
call \textit{cos-MDS} to visualize the neighbor distribution around an
agent. This method embeds the agent at the center and places its
immediate high-D neighbors around it.
The distance from each neighbor to the agent in the visualization represents their true distance in the high-D space. Neighbors' pairwise cosine distance, measured from the agent, approximates their true cosine distance in the high-D space.
The method effectively reveals the distances and distribution of neighbors around the agent, as well as the shape and topology of the empty space (fully enclosed, semi-enclosed, cluster boundaries, or anti-hub outliers). 

\textit{cos-MDS} is inspired by MDS to arrange the neighbors around the agent such that the shape and topology of their constellation is revealed in 2D. Similar to MDS, we normalize each agent-neighbor vector to the unit hypersphere, compute a pairwise cosine-distance matrix, perform eigenvalue decomposition, and use the top two eigenvalues and their eigenvectors for the low-dimensional embedding. Finally, we scale the low-D agent-neighbor vector to the true length in the original space (see 
App. \ref{appendix:cosmds} 
for a detailed description.)

The example shown in the ESC Editor (D) has neighbors uniformly
surrounding the configuration, and the distances to it are almost
equal. In essence this is a case where the empty space is fully
enclosed---a pocket in a high-D point cloud. Users can also change
the number of neighbors by clicking the \emph{+/-} button next to the
\emph{Neighbors} button in the control panel. This provides a more
comprehensive perspective to understand the empty space, \eg in a
hypersphere, within a paraboloid, or between hyperbolas. Detailed
examples can be found in App. \ref{appendix:cosmds}.

\subsection{Progress Tracker}
The Progress Tracker (D) keeps the user abreast of the achievements
made so far (F5). It has three main components (top to bottom): (1) a
Budget/Reward Display that captures the aggregated evaluation cost and
merit of the ESC exploration
so far, (2) a Training Status Display of the assistive DNN, and (3) a
Pareto Frontier Plot that shows the Pareto frontiers of existing
configurations (red) and
ESCs (gray) with respect to two user-chosen merit (target) variables. We now present these three displays in reverse order according to their use in practice. 

\textbf{Pareto Frontier Plot.}
The Pareto frontier is an effective mechanism for evaluating a set of
proposed configurations in the presence of multiple target
variables. To reduce visual complexity we currently restrict the
number of target variables to two. Our goal is to aid users in
recognizing trade-offs among the two target variables and identify
ESCs that can expand the frontier (push the envelope) at desirable
trade-off points on the curve.

We use the following color scheme in this plot: (1) existing configurations are colored blue, with their Pareto front connected by edges. The configurations defining these edges (and the edges themselves) are colored by their respective target variable values according to the red-toned color map to the left of the Overview Display and are represented as larger nodes there; (2) proposed configurations are colored gray, with their Pareto front connected by gray lines. Comparing the two curves facilitates an understanding of a candidate ESC's merits in the context what is known already and what the user's trade-off preferences are.

When the user modifies an ESC (say, in the PCP) its position in the
grey Pareto plot and frontier (and Overview Display) will update
accordingly. Then, upon verification, the ESC is added to the
``existing'' set, its Pareto point is colored blue, and the red Pareto
front updates.


\textbf{DNN Training Status Display.}
As mentioned, once the user loads the initial dataset, GapMiner trains an assistive DNN on the backend for ESC performance prediction. The DNN evaluates the performance of each proposed configuration 
before verification takes place. The testing error of the DNN is then shown in the DNN Training Status Display. We observe in (D) that there is a fairly large error in the initial stage (first part of the curve), hence at that stage user involvement is typically necessary to identify optimal configurations. The DNN is retrained whenever new configurations are added to the dataset and are verified.

The two horizontal lines in the display signify the user-specified DNN error tolerance. The DNN can be used to filter out poor configurations once it reaches below the first line, and it can be used for configuration optimization once it reaches below the second line. Details of the progressive search pipeline are introduced in Sec. \ref{sec:pipeline}.

\textbf{Budget/Reward Display.} There are also two other important objectives in practical applications, namely cost and benefit/reward. Maintaining a balanced optimization across both of these objectives is challenging but crucial. To gauge the progress of reward quantitatively, we incorporate the concept of \textit{Pareto dominance area} \cite{cao2015using} into the Progress Tracker (D) to monitor the Pareto front of the dataset. The Pareto dominance area is a well-established quality metric in multi-objective optimization that measures the volume of the objective space dominated by the Pareto front. As more ESCs are identified and verified, the Pareto front should extend further, dominating a larger area in the plot.

On the other hand, configuration verification can be expensive or time-consuming. For example, in the dataset we demonstrate, users had to purchase a configuration and run it for a week to obtain the real average throughput. In our application, thanks to CloudPhysics \cite{waldspurger2015efficient},  we could simulate a configuration quickly to get an approximated result. However, verification costs are unavoidable in many scenarios. To address this, we provide a budget counter in the Progress Tracker (D) to remind users of the aggregate cost of continued ESC verification.

We chose donut charts for both the Pareto Dominance-based Reward Display and the Budget Display due to their compactness and their ability to effectively convey proportional data at a glance. 

\section{DNN-Assisted  Configuration Search Pipeline}
\label{sec:pipeline}
While GapMiner effectively assists users in the search for optimal ESCs, manually identifying a large number of optimal ESCs can be tedious. To address this challenge, we propose a pipeline that trains an assistive DNN to eventually automate this process, positioning the analyst as a mentor to the DNN. This section provides a breakdown of the pipeline, as illustrated in
Fig. \ref{fig:workflow}. We begin by describing the various search methods we have implemented and then describe their role in the DNN-training process. 


\subsection{Configuration Search Methods}

In addition to the physics-inspired ESA, GapMiner integrates three more methods for investigating the empty space: random sampling, Pareto improvement, and a blank baseline. The blank baseline is a configuration with every variable set to 0.5, locating it at $[0,0]$ in the PCA map. This strategy provides no hint from the initial position, requiring users to rely solely on GapMiner interactions to optimize a configuration.

Random sampling, as the name suggests, samples configurations in a random manner. Pareto improvement, on the other hand, starts from the Pareto front of the existing set, allowing users to improve a configuration from a Pareto-optimal point. While Pareto improvement can be a good starting point for breakthroughs, it does not provide insights into the empty space. Empty-space search and random sampling are likely to find configurations better than those based on the Pareto front. Initially, we used another strategy, random walk, but experiments showed it was statistically similar to random sampling, so we did not include it in GapMiner.

All of these methods, including ESA, work as initial
configuration hints for users in the startup stage. Users can then
fine-tune the configurations proposed by these strategies to get
better configurations.
Additionally, users can also let the algorithm propose and
verify a batch of configurations. The batch size is determined by the
slider in the ESA Configurator in Fig. \ref{fig:teaser}(A). Once new configurations have
been verified, they will be added to the ``existing'' configuration set
and used to retrain the DNN. 

\begin{figure}[t]
    \centering
    \subfloat[]{\includegraphics[width=0.3\columnwidth]{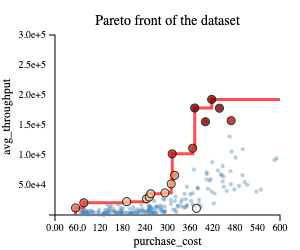}}
    \hfil
    \subfloat[]{\includegraphics[width=0.3\columnwidth]{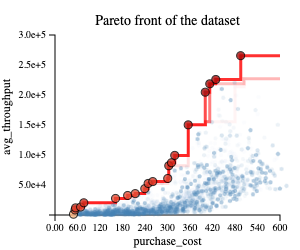}}
    \hfil
    \subfloat[]{\includegraphics[width=0.3\columnwidth]
    {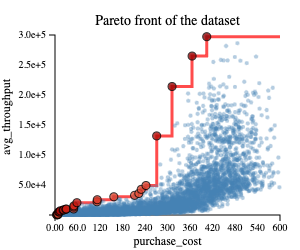}}
    \caption{An example Pareto front at the three key
      stages.  
      (a) The initial stage when the DNN has just started
      training. There are only few configurations in the ``existing''
      set, colored blue. (b) The front when the DNN has achieved
      $t_1$. The ``existing'' set has grown. (c)  The front when the
      DNN has achieved $t_2$. The ``existing'' set now covers much of
      the front's interior.}
    \label{fig:pf_milestones1}
    \vspace{-10pt}
\end{figure}


\subsection{Pipeline Throughout the DNN Training}
Fig. \ref{fig:pf_milestones1} shows the Pareto front development at the three pipeline stages. We now describe these in more detail. 

\textbf{Initial Stage.} In the early stage the dataset is still small and its empty space regions remain largely unexplored. With the data set still being insufficient to train a usable neural network, users can utilize GapMiner and its various exploration tools to identify empty spaces, propose new configurations, optimize them, and verify their effectiveness. In our studies with analysts we found that combining various configuration search strategies yields more diverse and higher-quality results to grow the dataset than sticking with only one search algorithm. Additionally, the visual feedback provided by the Progress Tracker incorporates elements of gamification. It not only indicates progress but also challenges users to find high-quality ESCs in a cost-effective manner.

\textbf{Developed Stages.} As the set of existing configurations grows, paying attention to the error plot in Fig. \ref{fig:teaser} (D) becomes crucial. As mentioned, it shows the average percentage error of the DNN along with the two user-set accuracy levels $t_1$ and $t_2$. Once the error is below $t_1$, the system will use the DNN to evaluate the proposed configurations, improving the quality of suggestions. Users can rely more on the system at this stage but still apply their domain knowledge to enhance results. Then, when the error drops below $t_2$, the system will go one step beyond, using gradient ascent
of the DNN to refine configurations further. The performance estimation now becomes accurate enough to find optimal configurations without user supervision. At this stage, the DNN and ESA can completely replace the user.

\section{Application Example}
\label{sec:application}
\begin{figure}[t]
    \centering
    \subfloat[]{\includegraphics[width=1\columnwidth]{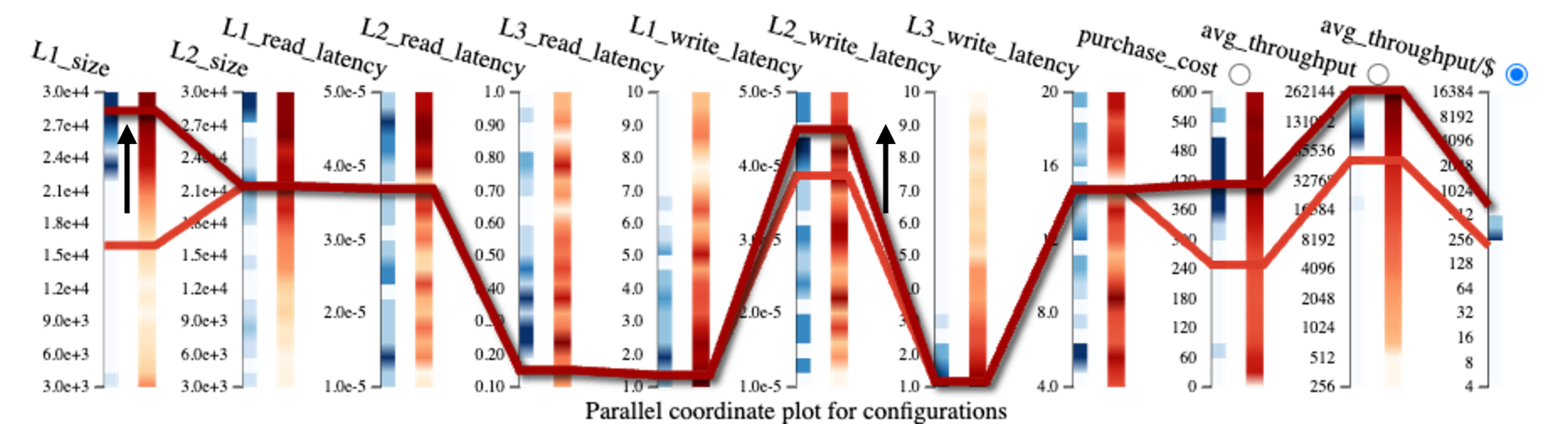}}
    \hfil
    \subfloat[]{\includegraphics[width=0.7\columnwidth]{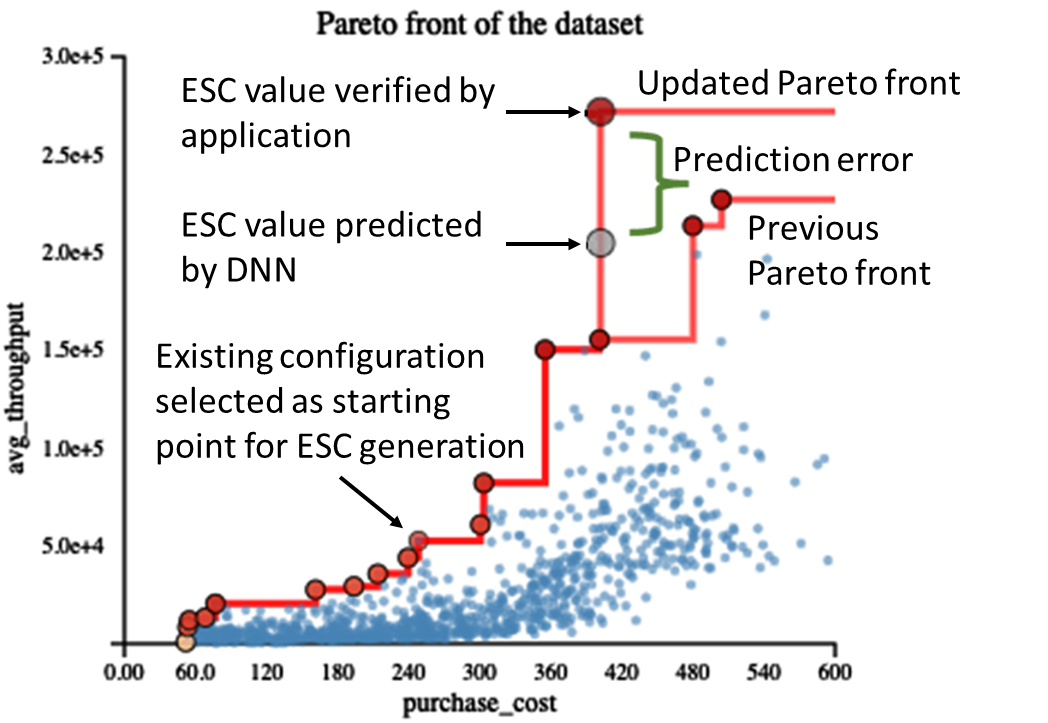}}
    \caption{Improvement of the Pareto front by interaction in the PCP followed by DNN estimation of the target values and subsequent verification in the domain application which revealed a lack of DNN training. (a) PCP showing the result of some user interactions; (b) annotated Pareto front.}
    \label{fig:appegs}
    \vspace{-10pt}
\end{figure}

To illustrate our pipeline, we present a use case  
in the application area of computer system optimization (see Sec. \ref{sec:termscolors} for an introduction to the data). Fig. \ref{fig:teaser} shows a snapshot taken of the interface during this session. 

%

We used a trace of a real workload run on actual machines, provided by
CloudPhysics \cite{waldspurger2015efficient}. We chose trace
\textit{w11}, which captures a week of virtual disk activity from a
production VMware environment. This trace file contains I/O requests
recorded over the week, which can then be replayed using either a
physical machine or a simulator of that machine. To save time and
energy, we used the simulator for all our experiments.

Given the size of the workload, evaluating a configuration is an expensive process. Additionally, there is a vast number of potential multi-tier configurations. Thus, conducting an exhaustive search of this space is infeasible, making the efficient identification of optimal configurations highly valuable to system administrators~\cite{hotstorage20mtcache, estro24peva}.

To begin, we collected a small number of configurations with the help of system experts.
We then invited a field expert to try GapMiner, ESA, and
the overall pipeline. Empirically, a configuration with good \textit{avg\_throughput}
is usually more expensive; thus the expert's goal was to find optimal
configurations with good performance at a lower price. 

The expert began by loading the dataset and setting the two DNN thresholds: $t_1 =
20\%$, $t_2 = 10\%$. Next, he selected  $\mathit{avg\_throughput}/\$$ as the
target variable since it captures a reasonable first compromise
between the two main objectives $\mathit{avg\_throughput}$ and
$\mathit{purchase\_cost}$. After bracketing the target variable he learned from
the range histogram (Fig. \ref{fig:teaser} (A)) that there were just a few
configurations in the best and worst value intervals.
%
Next the expert ran the ESA algorithm and then
checked the (higher) 8\textasciitilde10 interval of the ``existing''
group and the  6\textasciitilde8 and  8\textasciitilde10 groups from
the ``proposed'' configurations found by the ESA (see (Fig. \ref{fig:teaser}
(A)). This action revealed the former as a density plot and the latter
as scattered points in the Overview Display (Fig. \ref{fig:teaser} (B)). A
Pareto front, computed from the ``proposed'' configurations, appeared in
the Pareto front display (Fig. \ref{fig:teaser} (D)) in grey, with the
points defining the front being drawn as larger nodes in the Overview
Display.

The red curve in this Pareto front display shows the front of the
``existing'', verified configurations, and the configurations that
define it are colored by the selected target variable,
$\mathit{avg\_throughput}/\$$, according to the red-toned colormap to the left
of the Overview Display. The expert noticed that the grey front only
surpassed the red front in the high $\mathit{purchase\_cost}$ area but had poor
$\mathit{avg\_throughput}$ for low $\mathit{purchase\_cost}$
configurations. He decided
not to evaluate or verify any of the proposed ESCs but rather focused on
the red Pareto front in the low $\mathit{purchase\_cost}$ area. He found a
lower-performing configuration there, indicated by its orange, less
vibrant red color
(see Fig. \ref{fig:appegs}(b)). He first tried to improve it by moving it
to the dense region in the Overview Display where the high-performing
``existing'' configurations reside, but this strategy failed. To
investigate why, he examined the scree plot in Fig. \ref{fig:teaser} (A)
and noticed the slow growth of the aggregate curve. This essentially
means that the PCA map did not explain much of the data variance and
so moving configurations along loading vectors that pointed directly
to the dense region was not sufficient. Some short or nearly
orthogonal loading vectors like $\mathit{L1\_size}$ or
$\mathit{L1\_write\_latency}$
might also matter.

Looking for an alternative approach he turned to the density bars in the PCP to learn about the per-variable distributions. Fig. \ref{fig:appegs}(a) shows the PCP with the polyline of the Pareto configuration under consideration colored in orange, indicating its lower performance. And indeed, despite residing in a dense PCA region, this
configuration suffered from sub-optimal $\mathit{L1\_size}$ and
$\mathit{L1\_write\_latency}$
settings (it falls in less dense and lightly colored regions in the
density and correlation bars, respectively, for both variables). These
variables minimally contributed to the PCA space but apparently
significantly impacted the target variable outcomes.

The expert adjusted the configuration to dense intervals in both variables, as indicated by arrows in Fig. \ref{fig:appegs}(a). He then asked the DNN to 
estimate the value for $\mathit{avg\_throughput}$ and calculated other
target variables,
which showed significant improvement and expanded the Pareto front in
the upper region (grey node in Fig. \ref{fig:appegs}(b). Pleased, the
expert clicked the \emph{Verify} button to obtain the true
performance. After some time, the verification results showed that
this configuration was even better than the DNN prediction (red node
at the top of Fig. \ref{fig:appegs}(b)), and the dominance area increased
significantly.

While this was a positive development, the expert also realized that the DNN did not perform well in the upper region of the Pareto front (see the large prediction error in Fig. \ref{fig:appegs}(b)). More configurations in that range were needed to train the DNN effectively. To address this, the PCP interface enables users to select specific value intervals and run ESA exclusively within these intervals to generate ESC batches.

As the expert created more ESC batches and verified some of the
generated ESCs, the DNN eventually reached the $t_1$ threshold. At
this point, the expert was able to rely more on the DNN to assess the
quality of ESCs and he started to only focus on the ESCs located on
the proposed Pareto front for verification. Finally, once the DNN
surpassed $t_2$, the AI model completely took over the expert's
role. As shown in Fig. \ref{fig:pf_milestones1}, this three-stage pipeline
increased the dominance area from 0.27 to an impressive 0.56, more
than doubling it.

\section{User Study}
\label{sec:userstudy}

We conducted a user study to evaluate GapMiner with the same system
dataset (see 
App. \ref{appendix:userstudy}
for a detailed description of this experiment). We
asked 10 graduate students with CS backgrounds to perform
an optimal-configuration search from an initial stage that had 200
configurations
in the set. The users could make full use of GapMiner with all
search methods: ESA, random sampling, Pareto
improvement, and the blank baseline. The batch size for searching was
fixed to 50 for all users. They could click the \emph{search} button
multiple times, but they could click the \emph{verify} button only
5 times to get the true $\mathit{avg\_throughput}$. Once they believed
they had a
good configuration, they ran one of their alloted verifications. 
We calculated the
dominance area and updated it in the reward donut chart in Fig. \ref{fig:teaser}
(D)\cite{cao2015using} as the evaluation metric.

\subsection{Result Analysis}

\begin{figure}[tbp]
  \centering
  \includegraphics[width=0.9\columnwidth]{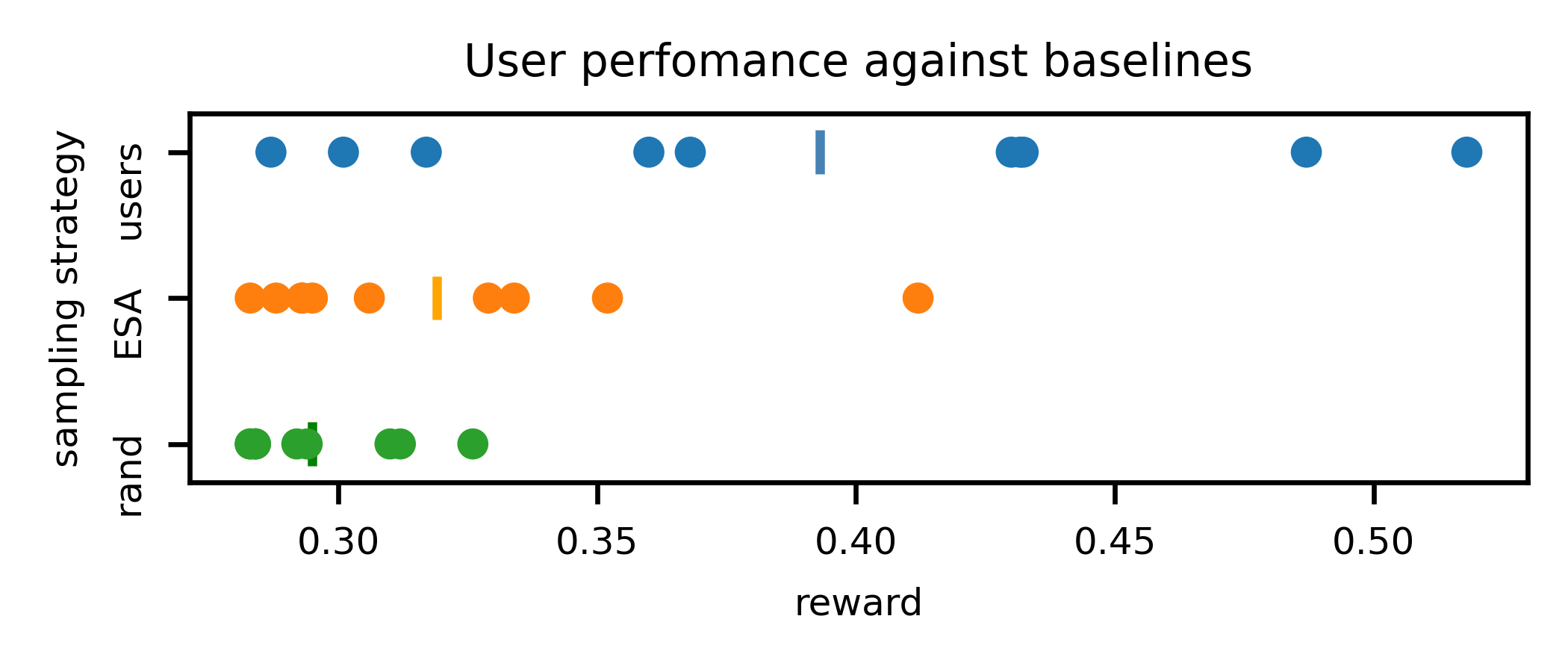}
    \vspace{-10 pt}
  	\caption{Rewards achieved by users, ESA, and random
          sampling. Each dot represents a search round (5
          verifications). The reward is defined by dominance area in the
          Pareto plot. The bar within each strip is the
          mean value.}
  \hfill%
  \label{fig:userstudy}
  \vspace{-15pt}
\end{figure}

We compared the GapMiner-aided performance of the users with the outcomes achieved with (1) ESA-based search only and (2) verifying the initial
random samples directly, in order to see whether GapMiner helped users with the
task. Noticing that users clicked \emph{search} three times on
average, we set the batch size to 150 for the two baselines (ESA and random) to match the
user behavior, calculated the Pareto front estimated by the naïve
neural network, and then verified the naïve Pareto front. We ran each
baseline 10 times to match the number of participants. We did not
choose the top five for verification in the baselines because we found
no difference between verifying the top five and the whole Pareto
front in the dominance area. Our results are given in
Fig. \ref{fig:userstudy}. We can see a clear advantage in user performance
compared to the two baselines. The average reward of ESA is
0.319 and the average reward of random sampling is 0.295, both of
which are much lower than our users’ average reward,
\textbf{0.393}. \textbf{7} users reached higher rewards than ESA and \textbf{9} reached higher rewards than random sampling.

We also did a between-subjects ANOVA with Tukey's Honestly
Significant Difference (HSD) to analyze
our data's statistical significance; the analysis is shown in Tab. \ref{tab:useranaly} in App. \ref{appendix:userstudy}. Cohen's $f$ from ANOVA was \textbf{0.85},
indicating a large effect size. We observe that GapMiner is significantly better than both baselines, while the outcomes between the two baselines do not differ statistically.


\subsection{System Usability}

We evaluated the usability of GapMiner with the popular questionnaire
System Usability Scale (SUS)\cite{sus}, which has 10 carefully
designed questions that evaluate the system in various
aspects. 
We distributed the SUS to the users after completion of the study and received 5
responses. Our score calculated from the questionnaire was \textbf{73}.
According to SUS
  score analyses from Bangor et al.~\cite{bangor2009determining} and
  Sauro et al.~\cite{sauro2011practical}, our system is ranked B (SUS
  score $>$ 72.6) and is thus considered good (SUS score $>$ 71.1).
Therefore, we believe that GapMiner has shown good usability in these
initial tests.

\subsection{Comparison Experiments}
\label{sec:compare}
In addition to testing user performance against algorithms in the
initial stage of the pipeline, we also did two
  experiments to compare how ESA outperforms random methods in
  developed stages with the help of a neural network. These
  experiments did not involve users, but employed the well-trained DNN
  to evaluate and fine-tune ESCs. In the standard pipeline described
  in Sec. \ref{sec:pipeline}, there is an evolutionary DNN working as a
  critic to determine promising ESCs, and a black box to collect true
  values. Based on the advice of field experts, we simplified the
  pipeline in the block-trace scenario by replacing the critic and the
  black box with a well-trained DNN. The DNN had an average percentage
  error of $7.9\%$, working as a surrogate model to simultaneously
  determine good ESCs and verify outcome values.

We employed two baselines in this section: random sampling (RS) and
random walk (RW).
Random sampling simply samples configurations in the data
space. Random walk further walks in a random direction for each of 400
steps, which is the same number we used in the ESA.

In the first experiment, we fast-forwarded to a developed stage that had 1,000 configurations in the dataset instead of iterating from an initial stage. 
In RS, we randomly sampled 1,500 configurations at once.
Starting from the same initial positions as used in RS,
we ran ESA and RW, and chose the best ones evaluated
by the DNN on each trajectory independently. We repeated the process
50 times to reduce result variance.
As before, we used the dominance area as the evaluation metric. The
average rewards of ESA, RS, and RW were
\textbf{0.450}, \textbf{0.409}, and \textbf{0.413}, respectively. ESA was statistically better than RS and RW,
and there was no significant difference between RS and RW. The effect
size $\eta^2$ was \textbf{0.3}, indicating that the
magnitude of the difference between the averages was \textbf{large}. The full 
statistical analysis is shown in Tab. \ref{tab:mid} in App. \ref{appendix:userstudy}.

In the second experiment, we fast-forwarded to a stage with 3,000 configurations, using the same surrogate DNN as before. We did the same
experiments as in the previous one, the difference being that we
used gradient ascent to optimize the best configuration for each
trajectory in ESA, RS, and RW. The average rewards were
\textbf{0.453}, \textbf{0.418}, and \textbf{0.422} respectively. There was a statistically significant difference between ESA and random baselines, but no difference between RS and
RW. $\eta^2$ was \textbf{0.3}, indicating a \textbf{large} level of effect
size. The full
statistical analysis is given in Tab. \ref{tab:final} in App. \ref{appendix:userstudy}.

All of these experiments and statistical analyses clearly show that our empty search algorithm is much better than random methods.

\section{Further Case Studies}
\label{sec:case}

In addition to the experiments in system research, we applied our work to an entirely different domain, wine investigation.
Furthermore, to show that empty-space search is also a valuable technique 
for other fields of research, we also demonstrate its potential
with adversarial learning in computer
vision and Reinforcement Learning (RL) in the MuJoCo environment. Our
case studies show the effectiveness of our ESA in a wide range of applications.

\subsection{Application in Wine Investigation}
\label{sec:wine_study}
We tested our work on the wine dataset \cite{sammari} to show generalizability to
domain-agnostic scenarios. Different from system verification, the wine
dataset was collected from the real world. It has 11 chemical
properties and a quality rating associated with each wine
instance. The 11 chemical properties include \{fixed, volatile,
citric\} acidity, residual sugar, chlorides, \{free, fixed\} sulfur
dioxide, density, pH, sulphates, and alcohol. The wine quality
ranges from 3 to 8.

\begin{figure}[t]
    \centering
    \subfloat[]{\includegraphics[width=0.5\columnwidth]{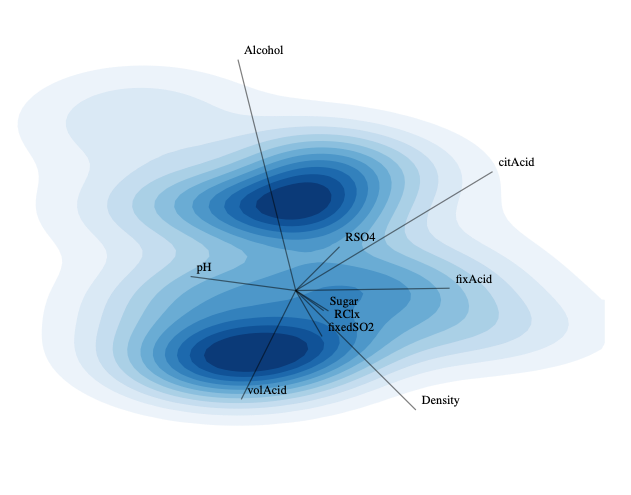}}
    \hfil
    \subfloat[]{\includegraphics[width=0.45\columnwidth]{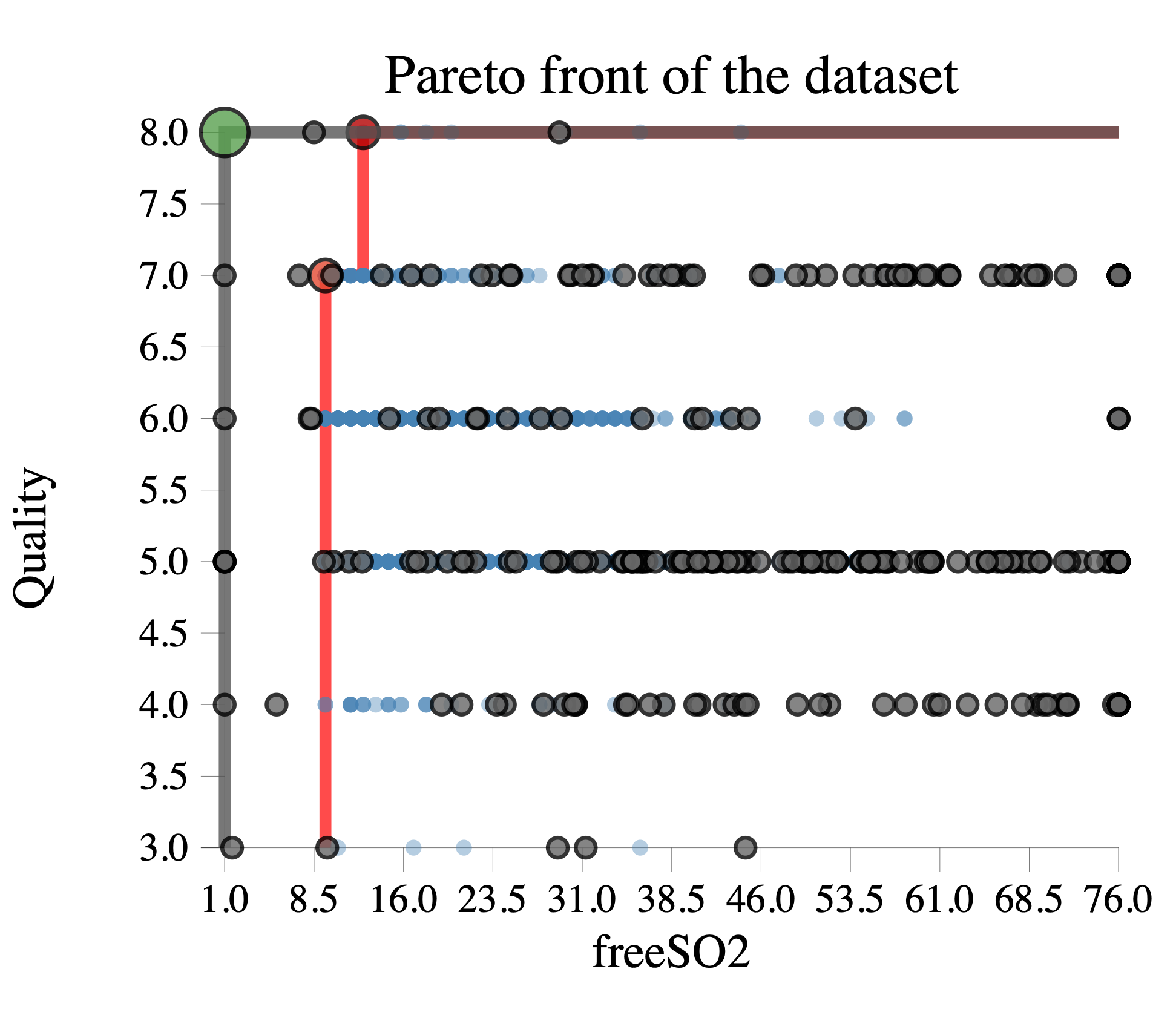}}
    \hfil
    \subfloat[]{\includegraphics[width=1\columnwidth]{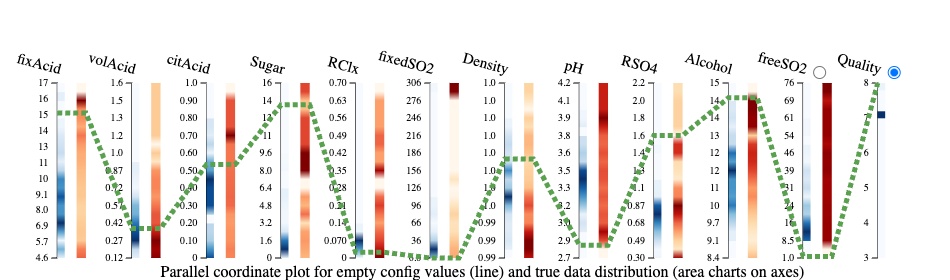}}
    \vspace{-5 pt}
    \caption{The key results of GapMiner on the wine dataset: (a)The PCA map showing the distribution of wine
          instances. The dark region on top represents wine instances
          with qualities 7 and 8, while the dark region below it corresponds to
          qualities 3 and 4; (b)ESA results in the Pareto plot. We plot the wine set
          as blue points and the associated Pareto front as a red
          line, while ESCs are gray points and the associated Pareto
          front is a gray line; (c) The full property values of the global optimal wine instance (highlighted in green in (b)).}
    \label{fig:winecase}
    \vspace{-15pt}
\end{figure}

Imagine there is a winemaker fermenting new wine. With quality being the
primary concern in mind, the winemaker also wants to minimize the free
sulfur dioxide due to potential health issues.
We can help the winemaker with GapMiner. First, the scree plot in
GapMiner reveals that the PCA space explains about 60\% of the
variance, which means that the corresponding region in the original
space is less variant given a dense region in the PCA space. Next, the
distribution of wine quality---3, 4, 7, and 8 in the PCA space (see
Fig. \ref{fig:winecase}(a))---illustrates that alcohol and citric acid point to
the high-quality region while volatile acidity and density point to
the low. Thus, given a wine instance, increasing alcohol and citric
acid while reducing volatile acidity and density is more likely to
improve the quality. Moreover, the PCP scented widgets indicate that
wine quality is independent from free sulfur dioxide. Thus, wine
instances in every quality can be fine-tuned to reduce free sulfur
dioxide, fitting a wide range of customers.

Looking at the histogram of outcome variable intervals, we noticed
that the dataset is severely imbalanced: almost 85\% of the wine
instances are ranked quality 5 or 6; only 1\% are ranked quality 3 or
8, which results in a low prediction accuracy. Therefore, we had to
augment the imbalanced quality groups. To do so, we first referred to
the PCP scented widgets to learn the property distributions of each
quality. For example, wine in quality 4 has higher volatile acidity
but lower citric acidity, while that in quality 7 is the
opposite. After finding the distributions, we constrained the search
range to a dense interval of each property by brushing on the scented
widgets in the PCP. Thus, ESA ran in a specific quality region.

Due to the difficulty of obtaining the quality of new wine instances
by ourselves, we made it the same as the nearest existing
neighbor. This simple but effective strategy is widely used in
imbalanced learning \cite{chawla2002smote}. Having identified over
6,000 ESCs, the DNN accuracy improved from 60\% to 80\% on quality
prediction.
On the other hand, free sulfur dioxide was integrated with other
properties during the empty-space search. Therefore, its value was
determined by ESA, which also reduced the DNN's mean-squared error from
0.018 to 0.002. During the pipeline execution, we were able to find
numerous wines with low free sulfur dioxide in each quality. We show the
results in Fig. \ref{fig:winecase} and 
App. \ref{appendix:wine}.

To summarize, our ESA, GapMiner, and the pipeline were able to balance the
dataset by augmentation, and improve DNN performance. Though we were
unable to verify these results in the field, our work could identify
numerous innovative wine instances predicted to have good quality or low
free sulfur dioxide.

\subsection{Empty-Space Images in Adversarial Learning}

Latent space has long been studied and has proven to be informative in
image generation \cite{karras2019style, abdal2019image2stylegan,
  li2023transforming}. In this section, we study the application of
GapMiner in the latent space of the MNIST dataset. We spanned the
latent space by using an AutoEncoder that transforms a 28*28 image to
a 4D point. Additionally, we trained a convolutional neural network
(CNN) to do a digit-recognition task and obtained an accuracy of $99.17\%$.

\begin{figure}[tbp]
  \centering
  \includegraphics[width=0.85\columnwidth]{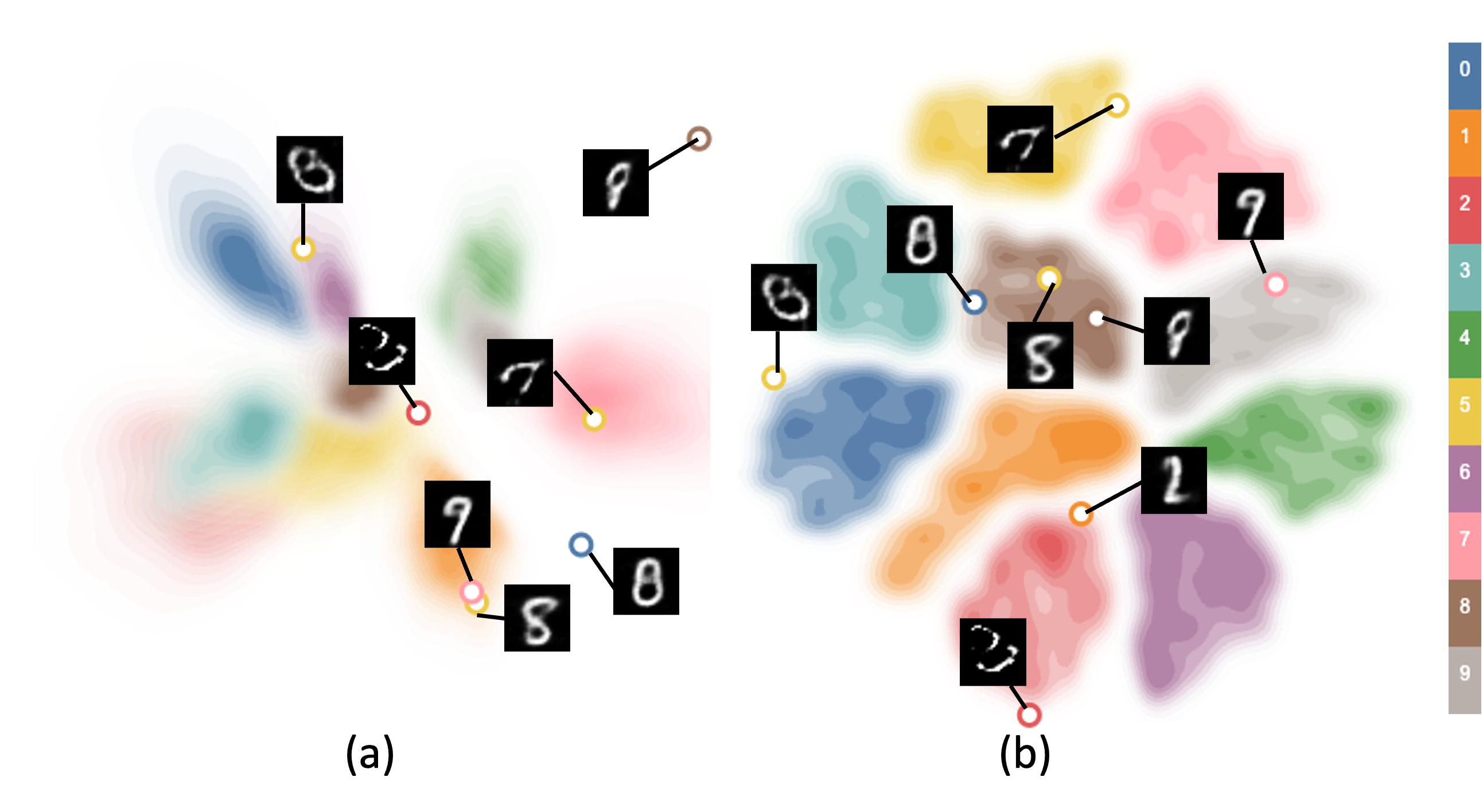}
  	\caption{Empty-space adversarial images from MNIST identified by
          ESA. We visualize the latent space by dimension reduction and highlight some of the adversarial points (compare a point's color with the colors of the clusters via the color legend on the right). 
          We show the PCA map at (a),
          which is directly from GapMiner, and the tSNE map at
          (b). All the images from MNIST are abstracted to
          contours. We can see clear clusters of digits.}
  \hfill%
  \label{fig:ES_adv}
  \vspace{-10 pt}
\end{figure}

In this experiment, we initialized agents randomly in the latent space
and ran ESA without momentum. Then we reconstructed images from
empty-space points. The results showed numerous meaningful empty-space
images that were falsely classified by the CNN. We show some of the
adversarial images in Fig. \ref{fig:ES_adv} and describe additional
details in App. \ref{appendix:adversarial}.
It is interesting to note that though these empty-space images in
Fig. \ref{fig:ES_adv}(b) were falsely predicted by the CNN, some of them
were mapped to the right cluster by tSNE.
While latent space has long been studied for image generation, our
work shows that there are many potential adversarial attacks within
it, which is an open research area.

\subsection{Empty-Space Search in RL}

RL research and its potential to exceed human performance have
attracted great attention. Specifically, Actor-Critic
\cite{konda1999actor} is an important RL algorithm that has been
widely used as the backbone in a broad range of fields, \eg in offline
meta-RL, network management, and trading \cite{li2024towards, chen2023distributional, demir2022statistical}. The
Actor-Critic algorithm trains a value network to evaluate the value of
an observation, and an independent policy network to take an action in
reaction to the observation via the feedback from the value network.
Here we employed a popular variation, Soft Actor Critic (SAC) \cite{haarnoja2018soft}, for empty-space study.

Mujoco\cite{todorov2012mujoco} is a physical simulation engine widely accepted for its robot control environments as RL benchmarks. We deployed one of the environments, AntDir, whose task is to control an ant robot to move in a given direction. The reward of AntDir is defined as the velocity towards the right direction, the higher the better. Observations are in a 27D continuous space and actions form an 8D continuous space.

In this case, we studied the possibility of getting a better policy from empty-parameter-space exploration. We trained a SAC agent on the AntDir environment for 200 iterations with a checkpoint saved at each iteration. Next, we extracted all the parameters
from the policy network in each checkpoint and converted them to a 22,160D point. There are a total of 200 points in the 22,160D parameter space.

\textbf{Empty-Space Policies.} 
We first visualized the parameter space using PCA and found a clear
training trajectory from a scratch policy to a well-trained one. (The
PCA results can be found in App. \ref{appendix:rl}.)
Therefore, we focused on the promising local space where the last
20 checkpoints were located, and searched empty space policies. For
each of the last 20 policies, the average position of 5 nearest
neighbors was taken as an agent's initial position, giving 20
agents. Then we ran ESA to propose candidate policies.


We evaluated those empty-space policies by running them on AntDir to get
their performance. Each policy ran 10 episodes, and we used the
average of the cumulative reward over an episode (also known as the
return) as our metric. 
The results are listed in Tab. \ref{tab:emptypolicy}. 
Our empty-space policy clearly
outperformed the training policies, not only showing ESA's utility in
searching for DNN parameters, but also unveiling a new way to train
neural networks that is much less computation-intensive.

\begin{table}[tb]
  \caption{
  Performance of training policies and empty-space policies
  }
  \label{tab:emptypolicy}
  \scriptsize%
  \centering%
  \begin{tabu}{%
  	  r%
  	  	*{7}{c}%
  	  	*{2}{r}%
  	}
  	\toprule
  	Policy & Max Return &  Avg Return   \\
  	\midrule
  	Training Policy & 319.20 & 272.29 $\pm$ 24.23  \\
  	Empty-Space Policy & 	\textbf{362.79} & \textbf{330.27 $\pm$ 9.08} \\
  	\bottomrule
   
  \end{tabu}%
\end{table}

\section{Discussion and Future Work}
\label{sec:discussion}
Empty-space exploration for mining new innovative configurations in
high-dimensional parameter spaces is a promising activity that has
long been overlooked. In this paper, we propose a novel empty space
search algorithm (ESA) to identify empty space configurations,
integrate it into a newly devised HITL visual analytics system,
GapMiner, and incorporate it all into a workflow that gradually
evolves the search process from HITL to AI-driven search. Our user
study demonstrates the effectiveness of GapMiner, supported by
comparison experiments showcasing its superiority over random and
ESA-only search. Additionally, several case studies illustrate the
technique's potential across diverse fields. We plan follow-up
research to find broader applications in security fields, \eg network
intrusion in cybersecurity \cite{chen2023real}.

The current initialization strategy of ESA
agents is random sampling, which is likely to miss important
regions. In future work, we plan to integrate subspace analysis to achieve more efficient agent deployment.
Also, our ESA assumes that variables are mutually
independent. However, in real-world scenarios there may be complex causal relationship among
the variables which our algorithm might fail to
uncover. On the other hand, the emerging large language models have
shown effectiveness in causal relationship inference
\cite{kiciman2023causal, zhang2023explainable}. In future
research, we plan to leverage the power of large language models to
incorporate causal information into the empty-space search process.  

Another aspect to improve is the progressive
neural network training process. Currently, our empty-space search focuses on optimal
configuration search, but it never speeds up neural network training. As a topic of future work, we plan to incorporate active learning to identify
ESCs that are both significant to network convergence and optimal.

\section*{Acknowledgments}
This work was partially funded by NSF grant CNS 1900706. We thank
M\'ario Antunes for helpful discussions.%


%



\printbibliography


 

\begin{IEEEbiography}[{\includegraphics[width=1in,height=1.25in,clip,keepaspectratio]{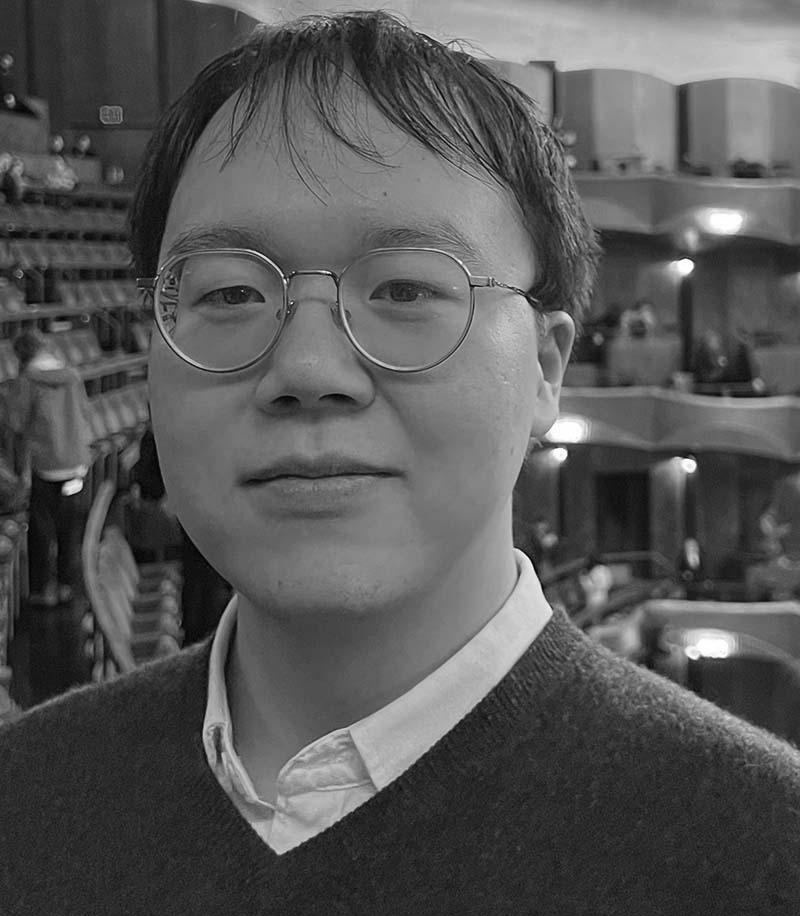}}]{Xinyu Zhang}
earned his B.E. in Computer Science at Shandong University, Taishan College in 2019. Currently, he is a Ph.D. candidate at Stony Brook University. His research interests include multivariate data analysis, scientific visualization, and reinforcement learning.
\end{IEEEbiography}

\begin{IEEEbiography}[{\includegraphics[width=1in,height=1.25in,clip,keepaspectratio]{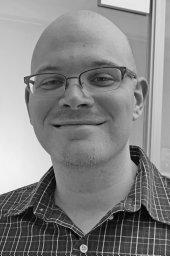}}]{Tyler Estro}
is a Ph.D. Candidate in Computer Science at Stony Brook University being advised by Professor Erez Zadok and a member of the File Systems and Storage Lab. Tyler’s primary research interests are multi-tier storage caching, disaggregated memory systems, and Compute Express Link (CXL) technology.
\end{IEEEbiography}

\begin{IEEEbiography}[{\includegraphics[width=1in,height=1.25in,clip,keepaspectratio]{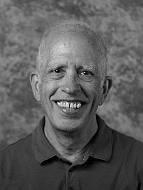}}]{Geoff Kuenning}
is an Emeritus Professor of Computer Science at Harvey Mudd College.  His research interests include storage systems, tracing, and performance measurement.  He is an Associate Editor of ACM Transactions on Storage, and the maintainer of the widely used SNIA IOTTA Trace Repository.
\end{IEEEbiography}

\begin{IEEEbiography}[{\includegraphics[width=1in,height=1.25in,clip,keepaspectratio]{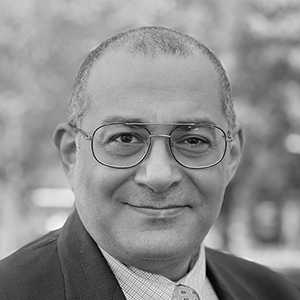}}]{Erez Zadok}
 is a CS Professor at Stony Brook University, where he directs the File Systems and Storage Lab (FSL).  His research interests include file systems and storage, operating systems, energy efficiency, performance and benchmarking, security and privacy, networking, and applied ML.  He received two SUNY Chancellor's Awards for Excellence and the NSF's CAREER Award.  He chaired several major conferences, served on steering committees, and is Editor-in-Chief of ACM Transactions on Storage.
\end{IEEEbiography}

\begin{IEEEbiography}[{\includegraphics[width=1in,height=1.25in,clip,keepaspectratio]{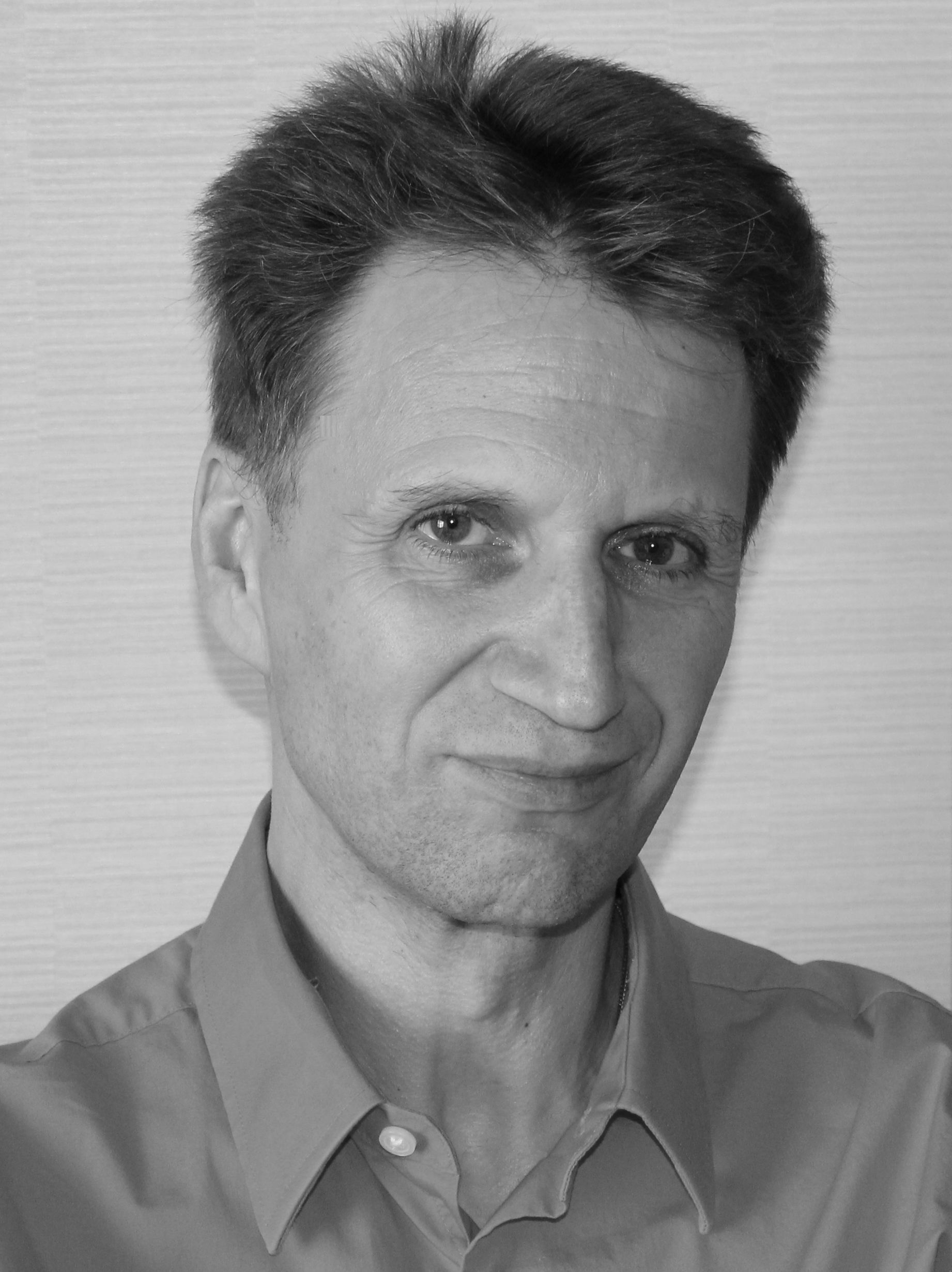}}]{Klaus Mueller}
is currently a professor of computer science at Stony Brook University and a senior scientist at Brookhaven National Lab. His research interests include explainable AI, visual analytics, data science, and medical imaging. He won the US NSF Early CAREER Award, the SUNY Chancellor’s Award for Excellence in Scholarship and Creative Activity, and the IEEE CS Meritorious Service Certificate. To date, his 300+ papers have been cited over 13,500 times. He was Editor-in-Chief of TVCG from 2019-22 and is a IEEE Fellow.
\end{IEEEbiography}



\vfill

\newpage

{\appendices
\section{Connection Between PCA Space and Original Space}
\label{appendix:pca_insights}
Given any multivariate configuration $V=[v_1, v_2, \dots v_N]^T$ in an
$N-$dimensional dataset, let the PCA transformation matrix be
\begin{gather*}
 M=\begin{bmatrix}
    m_{11} & m_{12}  & \dots  & m_{1N} \\
    m_{21} & m_{22}  & \dots  & m_{2N} \\
    \vdots & \vdots  & \ddots & \vdots \\
    m_{n1} & m_{n2}  & \dots  & m_{nN}
  \end{bmatrix}
\end{gather*}
The corresponding position $v = [v_1^{*}, v_2^{*}, \dots, v_n^{*}]^T$ in
$n-$dimensional PCA space is given by $v = MV$.
Changing the value of $V$ at the $k$th variable by $\delta v_k$ can be
written as $\delta V = [0,\dots \delta v_k, \dots, 0]$, and $V_{new} = V +
\delta V$.

Notice that the $i$th row of $M$ is the $i$th principal component in the
data space, $P_i = [m_{i1}, m_{i2}, \dots, m_{iN}]$, and the $j$th
column of $M$ is the loading vector of variable $j$: $L_j = [m_{1j},
  m_{2j}, \dots, m_{nj}]^T$. Therefore, the moving step $\delta v$ in
PCA space can be derived as follows:
\begin{equation}
    \label{eq:pcaexpo}
    \begin{aligned}
      v_{new} &= M  V_{new} \\  
      &= M  (V + \delta V) \\
      &= [P_1, P_2, \dots, P_n]^T  V + [L_1, L_2, \dots, L_N]  \delta V \\
      &= v + L_k  \delta v_k \\
    \Longrightarrow \delta v &= v_{new} - v \\
    &= L_k  \delta v_k
    \end{aligned}
\end{equation}
From Eq. \ref{eq:pcaexpo} we can conclude that the moving direction in
PCA space is determined only by the corresponding loading vector, and
the step size is determined by the value difference and the
loading vector's magnitude. Thus we can connect the original space in
the Parallel Coordinate Plot with the PCA space in the PCA map: changing the value on a PCP axis will lead
to a position update in PCA map as determined by Eq. \ref{eq:pcaexpo}.

\section{Agent-Centered Dimension Reduction}
\label{appendix:cosmds}
First we show how we embed a hypersphere. Given $n$ points $P = [P_1, P_2, \dots, P_{n}]^T$ on a
$d$-dimensional unit hypersphere, let $\Vec{P_i}$ be the vector from
$O$ to $P_i$.  Then we have
\begin{equation*}
    \begin{aligned}
        ||\Vec{P_i}|| &= 1 \forall P_i \in P \\
        \Vec{P_i} \Vec{P_j} &= ||\Vec{P_i}|| \; ||\Vec{P_j}||  cos\theta_{ij} \\
        &= cos\theta_{ij}
    \end{aligned}
\end{equation*}
Therefore, the product of any two points from the unit hypersphere is the cosine distance. We construct the cosine distance matrix as $M = PP^T$.

Then we perform the eigenvalue decomposition :
\begin{equation}
    \begin{aligned}
        M &= V\Lambda V^T \\
        &= V\Lambda^{\frac{1}{2}} \Lambda^{\frac{1}{2}} V^T \\
    \end{aligned}
    \label{eq:cos-mds-decomp}
\end{equation}
where $V$ is the eigenvector matrix and $\Lambda$ is the eigenvalue
diagonal matrix. Thus, we can conclude from Eq. \ref{eq:cos-mds-decomp}
that $P=V\Lambda^{\frac{1}{2}}$. Next we determine the $k$ largest
square-rooted eigenvalues $\Lambda^{*\frac{1}{2}}$ and corresponding
eigenvectors $V^*$. This yields $P^* = V^*\Lambda^{*\frac{1}{2}}$,
the low-dimensional embeddings of $P$ whose pairwise dot products fit
cosine distances in $P$.

\subsection{Nearest-Neighbor Visualization Algorithm}
\label{sec:dr_theory}

We show the process cos-MDS in
Alg. \ref{alg:neighbors}. In addition to dimension reduction, we also
scale each embedding vector to the same length as it has in the original
space, so that the distance from the embedding to the center
represents the true distance from the neighbor to the agent.

\subsection{Examples for Basic Geometric Structures}
We demonstrate the application of cos-MDS on three synthetic datasets: a 3D hyperboloid, a 4D paraboloid, and a 4D hypersphere. For each dataset, we approximately placed the agent at the center of the empty space and set the number of neighbors equal to the size of the dataset.

We generated the 3D hyperboloid dataset by uniformly sampling the 30 coordinates from $[-1, 1]$ on the $x$ and $y$ axis, respectively. We then calculated $z$ by 
$$
z = \pm 2*\sqrt{\frac{x^2}{0.04} + \frac{y^2}{0.04} + 1}
$$
There are 900 points in this dataset and we placed the agent at $(0, 0, 0)$. The synthetic dataset is directly plotted in Fig. \ref{fig:neighbors}(a). The corresponding result of cos-MDS can be found in Fig. \ref{fig:neighbors}(b).

The 4D paraboloid dataset was generated by uniformly sampling 10 points from $[-5, 5]$ on $x$, $y$, and $z$ axis independently, and the points on the paraboloid were defined by $(x, y, z, z^2)$. This dataset includes 1000 points. We set the agent at $(0, 0, 0, 20)$. The cos-MDS result on this dataset can be found in Fig. \ref{fig:neighbors}(c).

For the 4D hypersphere dataset, we first uniformly sampled 1,000 angles $(\psi, \theta, \phi)$, where $\psi \in [0, 2*\pi]$, $\theta \in [0, \pi]$, and $\phi \in [0, \pi]$. Then we calculated the points $(x, y, z, w)$ on the hypersphere by 
\begin{equation}
    \begin{aligned}
        x &= cos(\psi) sin(\theta) sin(\phi)\\
        y &= sin(\psi) sin(\theta) sin(\phi)\\
        z &= cos(\theta) sin(\phi)\\
        w &= cos(\phi)
    \end{aligned}
\end{equation}
This synthetic dataset contains 1,000 points as well. Again, we located the agent at (0, 0, 0, 0). The cos-MDS result can be found in Fig. \ref{fig:neighbors}(d).

From the results in Fig. \ref{fig:neighbors} we observe that cos-MDS visualizes a 3D hyperboloid (a) as a 2D hyperbola (b), transforms a 4D hypersphere into a 2D circle (d), and projects a 4D paraboloid onto the plane such that the distribution still approximates the pattern of a parabola (c). Our results indicate that this method successfully extracts the intrinsic features of the high-dimensional data distribution and intuitively visualizes them in the 2D plane, thus highlighting its great potential for empty space visualization.

\begin{algorithm}
\SetAlgoLined
 Pick $n$ nearest neighbors $[P_1, P_2, \dots, P_{n}]$ around the agent $\pi$ in $d$ dimensional space\;
 
 \For {i \dots $n$} {
    Let $\Vec{P_i}$ be the vector from $\pi$ to $P_i$\;
    $l_i$ = $||\Vec{P_i}||$\;
    $\Vec{P_i} = \Vec{P_i} / ||\Vec{P_i}||$\;
 }
 Calculate a local embedding $[\Vec{p_1}, \Vec{p_2}, \dots,
   \Vec{p_{n}}]$ from cos-MDS\;
 \For{i \dots {n}} {
    $\Vec{p_i} = l_i * \Vec{p_i} / ||\Vec{p_i}||$\;
 }
 return $[\Vec{p_1}, \Vec{p_2}, \dots, \Vec{p_{n}}]$ as neighbor plot results.
 \caption{Dimension Reduction for Cosine Distance}
 \label{alg:neighbors}
\end{algorithm}

\begin{figure}[t]
    \centering
    \subfloat[]{\includegraphics[width=0.45\columnwidth]{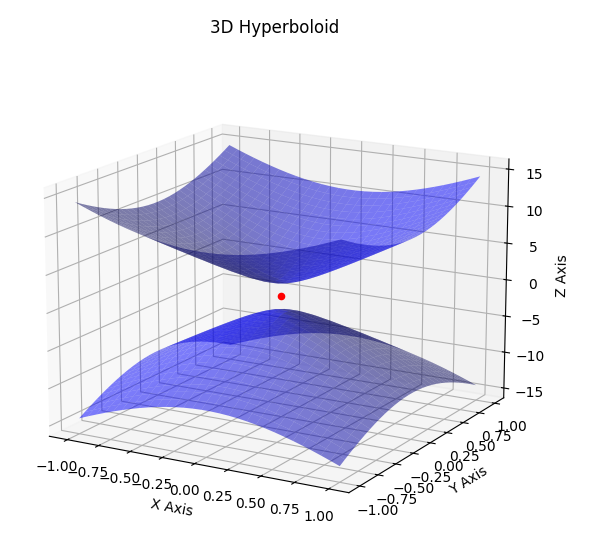}}
    \hfil
    \subfloat[]{\includegraphics[width=0.45\columnwidth]{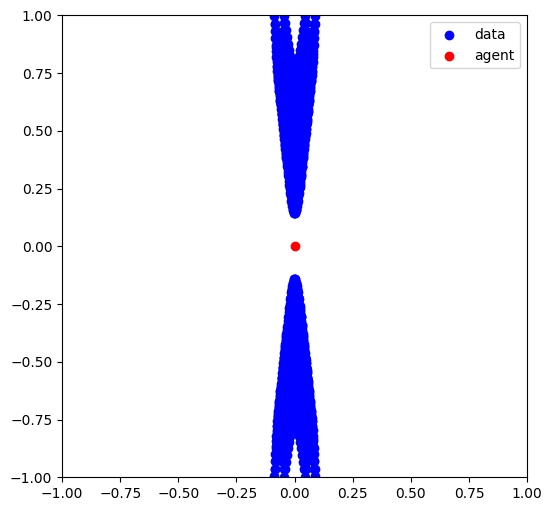}}
    \hfil
    \subfloat[]{\includegraphics[width=0.45\columnwidth]{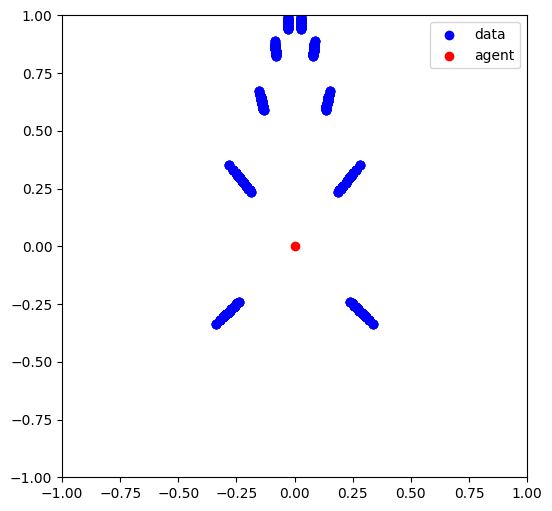}}
    \hfil
    \subfloat[]{\includegraphics[width=0.45\columnwidth]{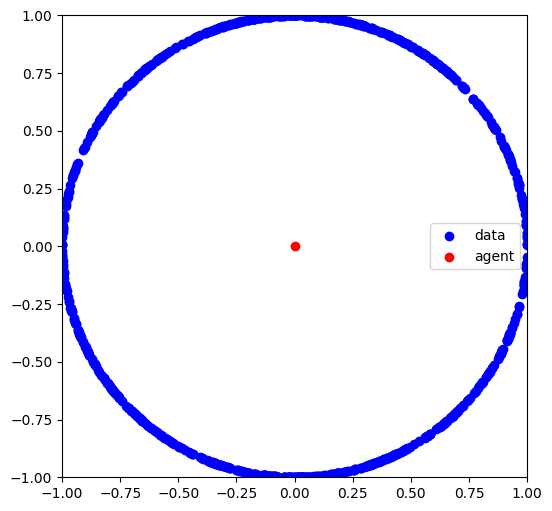}}
    \vspace{-5 pt}
    \caption{Examples of cos-MDS on 3 synthetic datasets. The blue points are the synthetic dataset and the red point is the center agent. (a) The 3D hyperboloid dataset; (b) cos-MDS result on the 3D hyperboloid dataset; (c) cos-MDS result on the 4D paraboloid dataset; (d) cos-MDS result on the 4D hypersphere dataset}
    \label{fig:neighbors}
    \vspace{-15pt}
\end{figure}

\section{Comparing ESA and Delaunay Triangulation}
\label{appendix:ESA-Delaunay}

\begin{figure}[t]
    \centering
    \subfloat[]{\includegraphics[width=0.45\columnwidth]{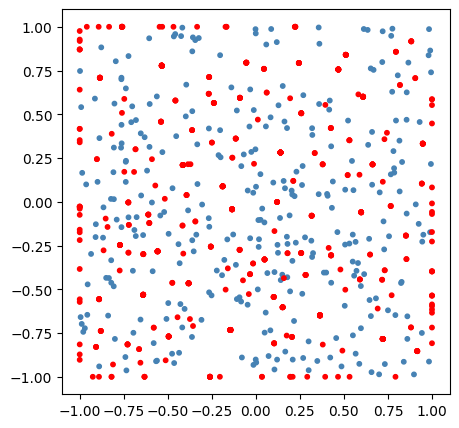}}
    \hfil
    \subfloat[]{\includegraphics[width=0.45\columnwidth]{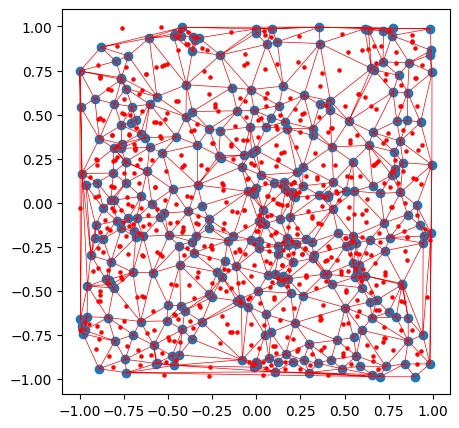}}
    \caption{Given 300 random samples (blue) and 600 random agents (red) in 2D space, we show the outcomes of empty space search using (a) our empty space search algorithm (ESA) and (b) Delaunay Triangulation (DT).}
    \vspace{-15pt}
    \label{fig:ESA-points1}
\end{figure}



Our investigation into high-dimensional Delaunay triangulation
acceleration and approximation revealed a scarcity of research
addressing the exponential complexity increase with dimensionality. Hence, we conducted some experiments to study this topic further. Our empirical findings underscore the computational and storage
impracticality of these types of approaches for our purposes, even for
datasets as small as 1,000 points, when the number of dimensions
exceeds 7. Fig. \ref{fig:ESA-points1}(a) and Fig.\ref{fig:ESA-points1}(b) show the empty space search results from our search algorithm (ESA) and from Delaunay Triangulation (DT) in the 2D case, respectively. While both methods identify all the empty spaces, DT requires many more triangles compared to the number of agents used in our ESA. It is this abundance (and complexity) of geometric primitives that limits DT's scalability to higher dimensions. 

\section{User-Study Results}
\label{appendix:userstudy}

The user-study data including user performance, ESA performance, and random
sampling performance are listed in Tab. \ref{tab:usdata}. The parameters
used in this study and comparison experiments were $k=9$,
$\sigma=$ mean distance of $k$ neighbors, $n=400$,
$\alpha=0.001$, $j=10$, $\delta=10^{-7}$, and $\gamma=0.9$, respectively.

\begin{table*}[tb]
  \caption{Rewards of each round of our user study (top), ESA (middle) and random sampling (below).
  }
  \label{tab:usdata}
  \scriptsize%
  \centering%
  \begin{tabu}{%
  	  r%
  	  	*{7}{c}%
  	  	*{2}{r}%
  	}
  	\toprule
  	User1 &  User2	 &   User3	 &   User4 & User5 & User6 & User7 & User8 & User9 & User10  \\
  	\midrule
  	0.487 & 0.4323 & 0.4317 & 0.36	 &  0.301	 & 0.368 & 0.43 & 0.518 & 0.287 & 0.317 \\
  	\bottomrule
   
  \end{tabu}%

  \begin{tabu}{%
  	  r%
  	  	*{7}{c}%
  	  	*{2}{r}%
  	}
  	\toprule
  	ESA1 & ESA2 &  ESA3	 &   ESA4	 &   ESA5 & ESA6 & ESA7 & ESA8 & ESA9 & ESA10 \\
  	\midrule
  	0.288 & 0.329 & 0.295 & 0.306 & 0.283 & 0.334 & 0.293 & 0.412 & 0.293 & 0.352  \\
  	\bottomrule
   
  \end{tabu}%

    \begin{tabu}{%
  	  r%
  	  	*{7}{c}%
  	  	*{2}{r}%
  	}
  	\toprule
  	Rand1 & Rand2 &  Rand3	 &   Rand4	 &   Rand5 & Rand6 & Rand7 & Rand8 & Rand9 & Rand10 \\
  	\midrule
  	0.31 & 0.326 & 0.283 & 0.283 & 0.284 & 0.312 & 0.284 & 0.292 & 0.294 & 0.284 \\
  	\bottomrule
   
  \end{tabu}%
\end{table*}

\subsection{ANOVA of User Study}
The ANOVA of user study can be found in Tab. \ref{tab:useranaly}.

\begin{table}[tb]
  \caption{The between-subjects ANOVA (above) and Tukey HSD (below) of user performance with GapMiner (x1)
    against the two baselines ESA-only (x2) and random samples-only (x3).   
  }
  \label{tab:useranaly}
  \scriptsize%
  \centering%
  \begin{tabularx}{\columnwidth}{%
  	  r%
  	  	*{7}{c}%
  	  	*{2}{r}%
  	}
  	\toprule
  	Source & DF &  SS	 &   Mean Square	 &   F Statistic &   P-value   \\
  	\midrule
  	Groups & 2 & 0.052 & 0.026	 &  9.746	 & $<$0.001  \\
  	Error & 27	 & 0.073	 & 0.003 &  &  \\
  	Total & 29	 &  0.125	 & 0.004 &  &  & \\
  	\bottomrule
   
  \end{tabularx}%

  \begin{tabularx}{\columnwidth}{%
  	  r%
  	  	*{7}{c}%
  	  	*{2}{r}%
  	}
  	\toprule
  	Pair & Diff &  SE	 &   Q	 &   CI Bounds & CM & P-value \\
  	\midrule
  	x1-x2 & 0.075	 & 0.016	 & 4.555	 &  0.017\textasciitilde0.132	& 0.058	& 0.009  \\
  	x1-x3 & 0.098	 & 0.016	 & 5.976 &  0.041\textasciitilde0.156	& 0.058 &	$<$0.001&  \\
  	x2-x3 & 0.023	 &  0.016 & 1.421	 & -0.034\textasciitilde0.081 & 0.058	& 0.580 \\
  	\bottomrule
   
  \end{tabularx}%
\end{table}

\subsection{System Usability Scale}
The SUS questionnaire included the 10 questions listed below:
\begin{itemize}
    \item I think that I would like to use this system frequently.
    \item I found the system unnecessarily complex.
    \item I thought the system was easy to use.
    \item I think that I would need the support of a technical person to be able to use this system.
    \item I found the various functions in this system were well integrated.
    \item I thought there was too much inconsistency in this system.
    \item I would imagine that most people would learn to use this system very quickly.
    \item I found the system very cumbersome to use.
    \item I felt very confident using the system.
    \item I needed to learn a lot of things before I could get going with this system
\end{itemize}
Each question was ranked from 1 (strongly disagree) to 5 (strongly
agree). The final score was calculated by summing the normalized scores
and multiplying them by 2.5 to convert the original score of 0-40
to 0-100. 5 users responded to our questionnaire. The results are given in Tab. \ref{tab:susdata}.
\begin{table}[tb]
  \caption{Responses from 5 users. Q1--Q10 refer to 10 questions;
    R1--R5 refer to user responses. NS in the last row is the
    normalized score over 5 responses.
  }
  \label{tab:susdata}
  \scriptsize%
  \centering%
  \begin{tabu}{%
  	  r%
  	  	*{9}{c}%
  	  	*{2}{r}%
  	}
  	\toprule
  	Resp & Q1 &  Q2	 &   Q3	 &   Q4 & Q5 & Q6 & Q7 & Q8 & Q9 & Q10  \\
  	\midrule
  	R1 & 3 & 2 &	4 &	2 &	4 &	2 &	4 &	2 &	4 &	3 \\
        R2 & 5	& 1 &	4 &	4 &	5 &	1 &	2 &	2 &	4 &	3\\
        R3 & 5 &	2 &	3 &	2 &	3 &	1 &	4 &	2 &	5 &	4\\
        R4 & 2 &	3 &	4 &	4 &	5 &	1 &	5 &	3 &	4 &	4\\
        R5 & 4 &	1 &	4 &	2 &	5 &	1 &	5 &	1 &	5 &	3\\
        NS & 2.8 &	3.2 &	2.8 &	2.2 &	3.4 &	3.8 &	3 &	3 &	3.4 &	1.6\\
  	\bottomrule
   
  \end{tabu}%

\end{table}

\subsection{Comparison Experiment Data}

Fig. \ref{fig:comparison_data} presents detailed results achieved by ESA, random sampling and random walk.

\begin{figure}[t]
    \centering
    \subfloat[]{\includegraphics[width=\columnwidth]{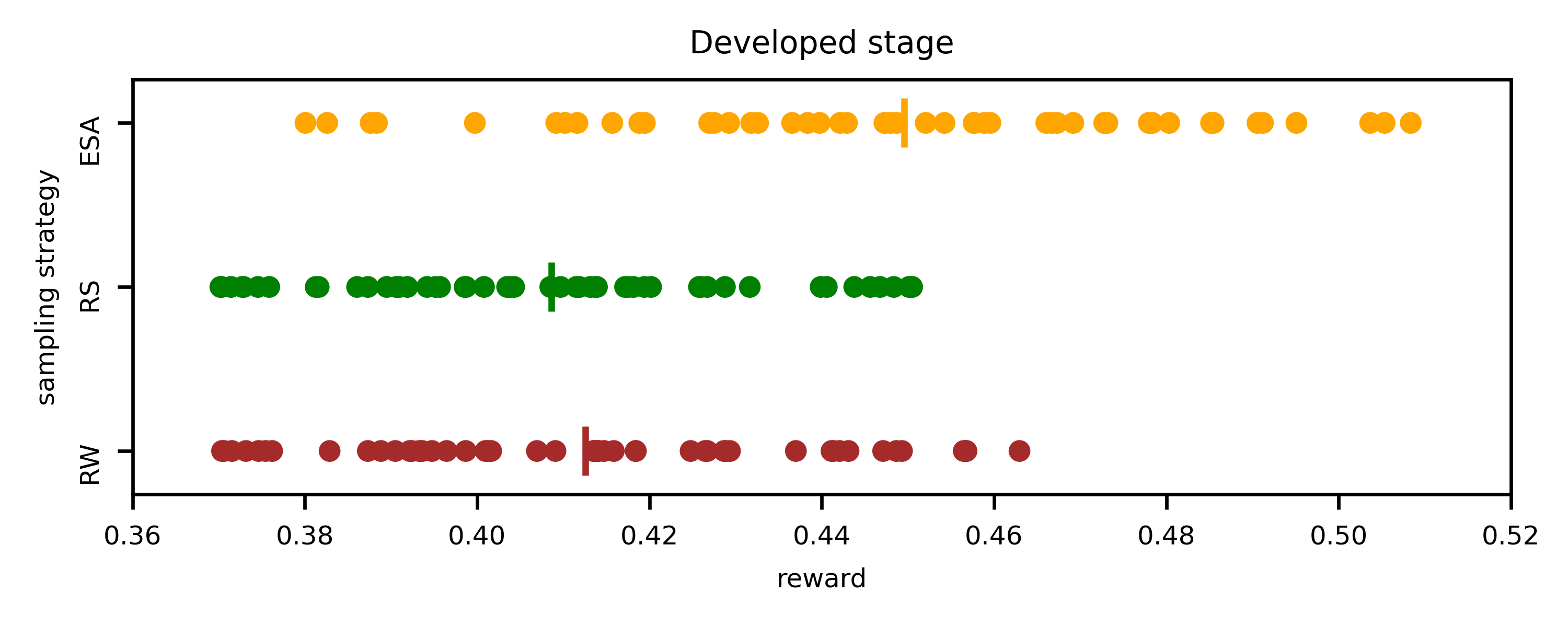}}
    \hfil
    \subfloat[]{\includegraphics[width=\columnwidth]{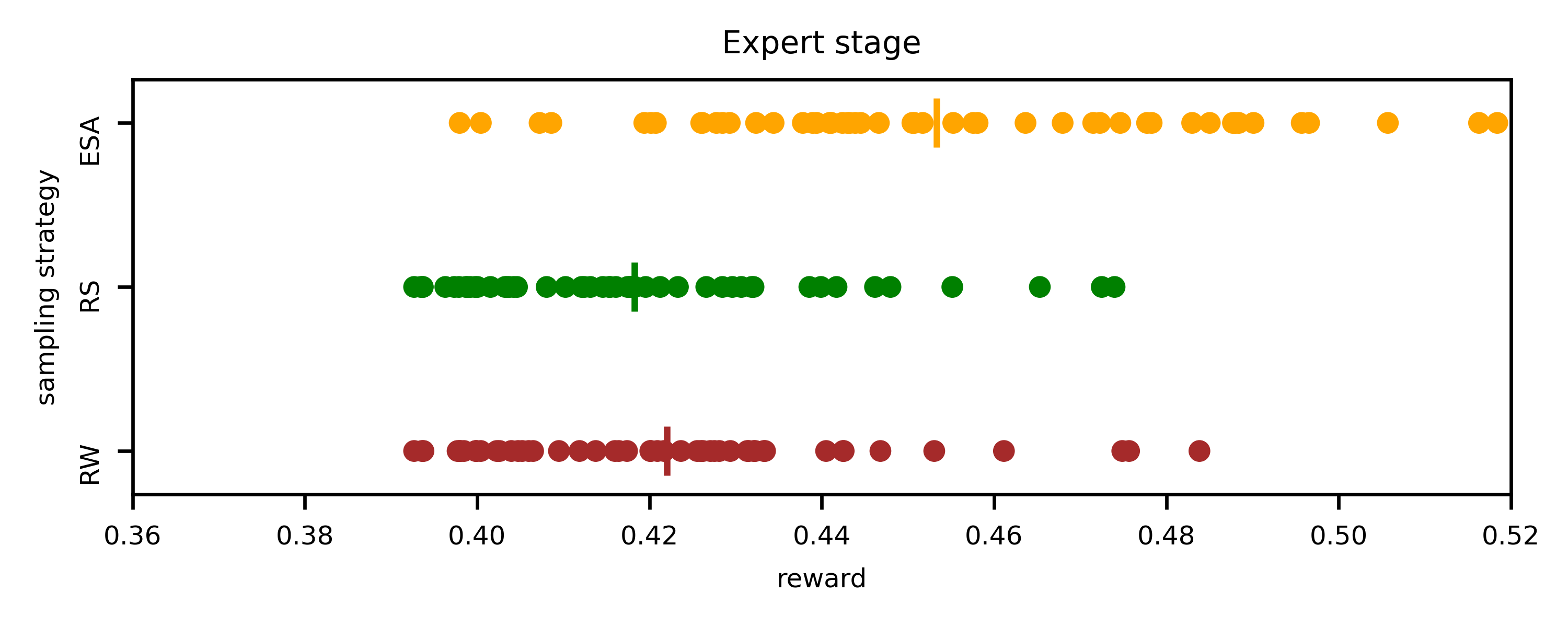}}
    \caption{Comparison experiment: (a) shows the rewards found by ESA,
    random walk, and random sampling at the development stage;
    (b) shows the rewards found by the three methods at
    the expert
    stage. In both figures, each dot represents a search process with
    1,500 random agents involved. Each method in both stages has 50
    search processes.}
    \vspace{-15pt}
    \label{fig:comparison_data}
\end{figure}

\subsection{ANOVA of Comparison Experiments}
The ANVOA of developed stage and expert stage can be found in Tab. \ref{tab:mid} and Tab. \ref{tab:final} respectively.

\begin{table}[tb]
  \caption{The within-subjects ANOVA (above) and Tukey HSD (below) of ESA (x1), RS (x2), and RW (x3) at a developed stage. Subjects are the initial positions of agents, Treatments are the empty space search methods.
  }
  \label{tab:mid}
  \scriptsize%
  \centering%
  \begin{tabularx}{\columnwidth}{%
  	  r%
  	  	*{7}{c}%
  	  	*{2}{r}%
  	}
  	\toprule
  	Source & DF &  SS	 &   MS	 &   F Statistic &   P-value   \\
  	\midrule
        Between Subjects & 49 & 	0.042 & 	0.001 & 1.115 (49,98) & 0.320 \\
  	Between Treatments & 2 & 0.051 & 0.026	 &  33.224 (2,98)	 & $<\mathbf{10^{-6}}$  \\
  	Error & 98	 & 0.075	 & 0.001 &  &  \\
  	Total & 149	 &  0.168	 & 0.001 &  &  & \\
  	\bottomrule
   
  \end{tabularx}%

  \begin{tabularx}{\columnwidth}{%
  	  r%
  	  	*{7}{c}%
  	  	*{2}{r}%
  	}
  	\toprule
  	Pair & Diff &  SE	 &   Q	 &   CI Bounds & CM & P-value \\
  	\midrule
  	x1-x2 & 0.041	 & 0.004	 & 10.250	 &  0.028\textasciitilde 0.054	& 0.013	& $<\mathbf{10^{-6}}$  \\
  	x1-x3 & 0.037	 & 0.004	 & 9.272 &  0.024\textasciitilde 0.050	& 0.013 &	$<\mathbf{10^{-6}}$&  \\
  	x2-x3 & 0.004	 &  0.004 & 0.979	 & -0.009\textasciitilde 0.017 & 0.013	& 0.769 \\
  	\bottomrule
   
  \end{tabularx}%
\end{table}

\begin{table}[tb]
  \caption{The within-subjects ANOVA (above) and Tukey HSD (below) of ESA (x1), RS (x2), and RW (x3) at an expert stage. Subjects are the initial positions of agents, Treatments are the empty space search methods.
  }
  \label{tab:final}
  \scriptsize%
  \centering%
  \begin{tabularx}{\columnwidth}{%
  	  r%
  	  	*{7}{c}%
  	  	*{2}{r}%
  	}
  	\toprule
  	Source & DF &  SS	 &   MS	 &   F Statistic &   P-value   \\
  	\midrule
        Between Subjects & 49 & 0.033 & 0.001 & 1.201 (49,98) & 0.220\\
  	Between Treatments & 2 & 0.037 & 0.019	 &  	33.233 (2,98)	 & $\mathbf{<10^{-6}}$  \\
  	Error & 98	 & 0.055	 & 0.001 &  &  \\
  	Total & 149	 &  0.125	 & 0.001 &  &  & \\
  	\bottomrule
   
  \end{tabularx}%

  \begin{tabularx}{\columnwidth}{%
  	  r%
  	  	*{7}{c}%
  	  	*{2}{r}%
  	}
  	\toprule
  	Pair & Diff &  SE	 &   Q	 &   CI Bounds & CM & P-value \\
  	\midrule
  	x1-x2 & 0.035	 & 0.003	 & 10.161	 &  0.024\textasciitilde 0.047	& 0.012	& $\mathbf{<10^{-6}}$  \\
  	x1-x3 & 0.031	 & 0.003	 & 9.082 &  0.020\textasciitilde 0.043	& 0.012 &	$\mathbf{<10^{-6}}$&  \\
  	x2-x3 & 0.004	 &  0.003 & 1.079	 & -0.008\textasciitilde 0.015 & 0.012	& 0.726 \\
  	\bottomrule
   
  \end{tabularx}%
\end{table}

\section{Extrapolation Behavior of Our ESA-Based Search}
Our ESA focuses on identifying empty spaces within the data distribution. The repulsive component prevents it from straying too far from the existing distribution. However, empty spaces outside the current distribution can be meaningful and essential. This experiment demonstrates ESA's ability to search for these empty spaces beyond the current distribution.

We used the distribution of the wine dataset for demonstration; we ignored the meaning of each variable and removed the target variable to focus solely on the distribution of the dataset in each dimension. This marked the dataset as the initial distribution. We ran ESA for six iterations, selecting 300 data items from the dataset with the largest average distance to their 8 nearest neighbors as the initial agent position. ESA search was conducted, and the results were added to the dataset. After each iteration, we measured the distance of the ESA results from the initial distribution and created a histogram. The histograms, shown in Fig. \ref{fig:extrapolation}, indicate that the configurations found by ESA progressively move further from the initial distribution. This demonstrated ESA's ability to explore empty spaces beyond an initial data distribution.

\begin{figure}[t]
    \centering
    \subfloat[]{\includegraphics[width=0.45\columnwidth]{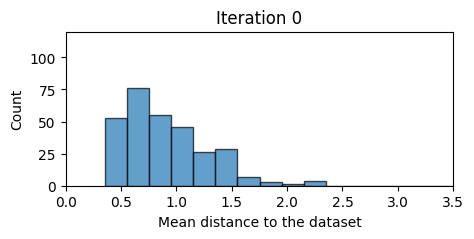}}
    \hfil
    \subfloat[]{\includegraphics[width=0.45\columnwidth]{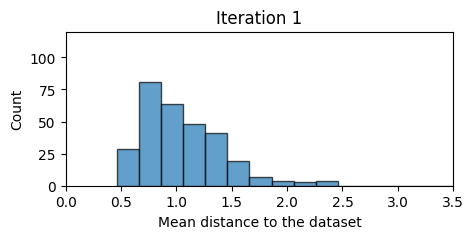}}
    \hfil
    \subfloat[]{\includegraphics[width=0.45\columnwidth]{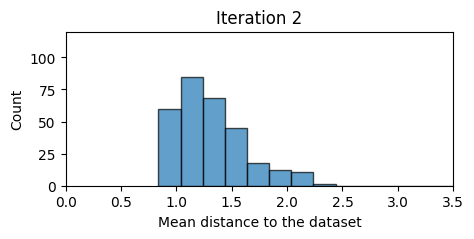}}
    \hfil
    \subfloat[]{\includegraphics[width=0.45\columnwidth]{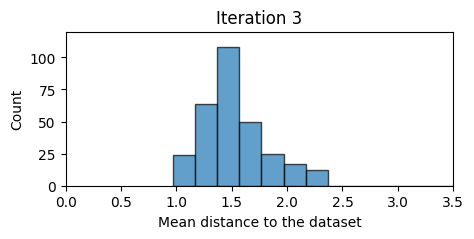}}
    \hfil
    \subfloat[]{\includegraphics[width=0.45\columnwidth]{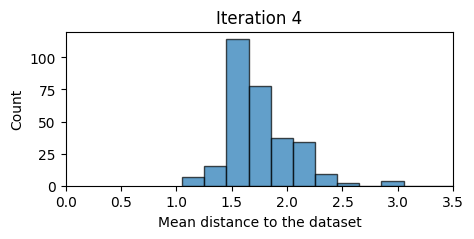}}
    \hfil
    \subfloat[]{\includegraphics[width=0.45\columnwidth]{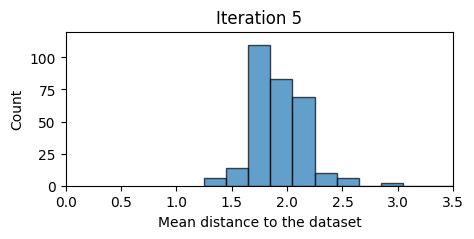}}
    \caption{Histograms of the configurations identified by the ESA with regards to the initial distribution as a function of iteration. We observe that the ESA results progressively move further from the initial distribution, demonstrating ESA's ability to explore empty spaces beyond an initial data distribution.}
    \vspace{-15pt}
    \label{fig:extrapolation}
\end{figure}


\section{Details for the Application Cases}
In  this section we present more detail on the various application examples shown in the paper. 

\begin{figure}[t]
    \centering
    \subfloat[]{\includegraphics[width=\columnwidth]{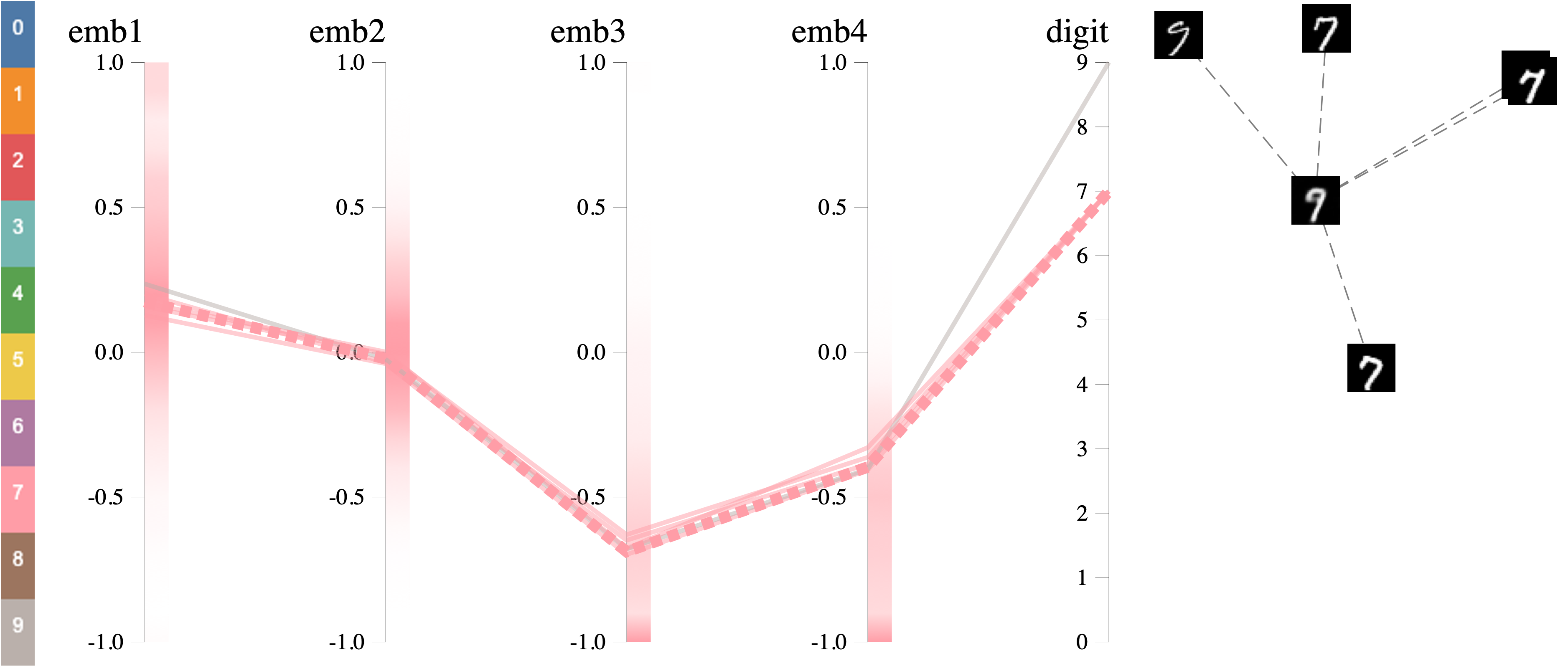}}
    \hfil
    \subfloat[]{\includegraphics[width=\columnwidth]{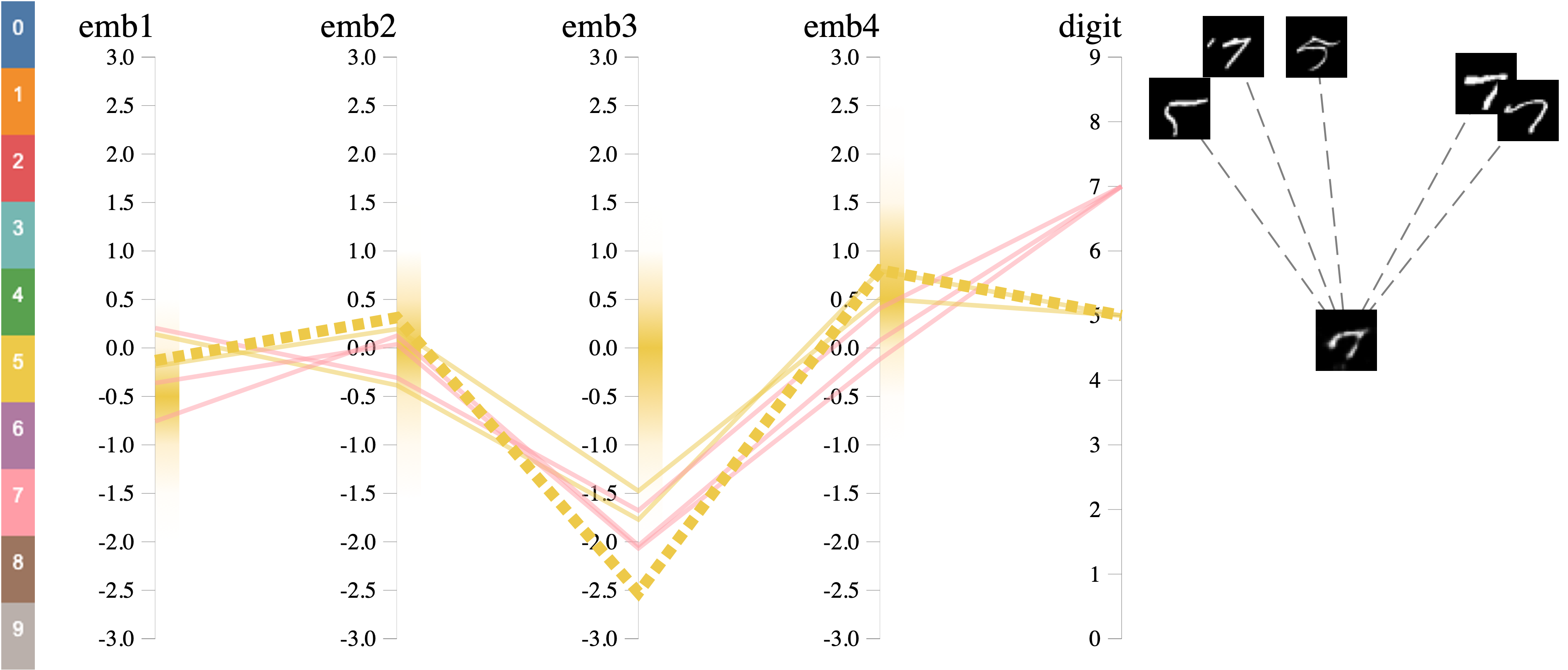}}
    \caption{Some key results from GapMiner. Fig. \ref{fig:ES_adv}
    shows the corresponding PCA map. The leftmost color bar is the legend, where
    each color represents a digit. The dashed line in the PCP is the
    empty-space image and the solid lines are its neighbors from MNIST. The line
    color refers to the true label of images from MNIST or the
    predicted label of the empty-space image. On the right is the neighbor plot,
    where the center is the empty-space image (dashed line in the PCP) and
    around it are the MNIST neighbors (solid lines in the PCP).}
    \vspace{-15pt}
    \label{fig:casemnist}
\end{figure}


\begin{figure}[tbp]
  \centering
  \vspace{5pt}
  \includegraphics[width=0.85\columnwidth]{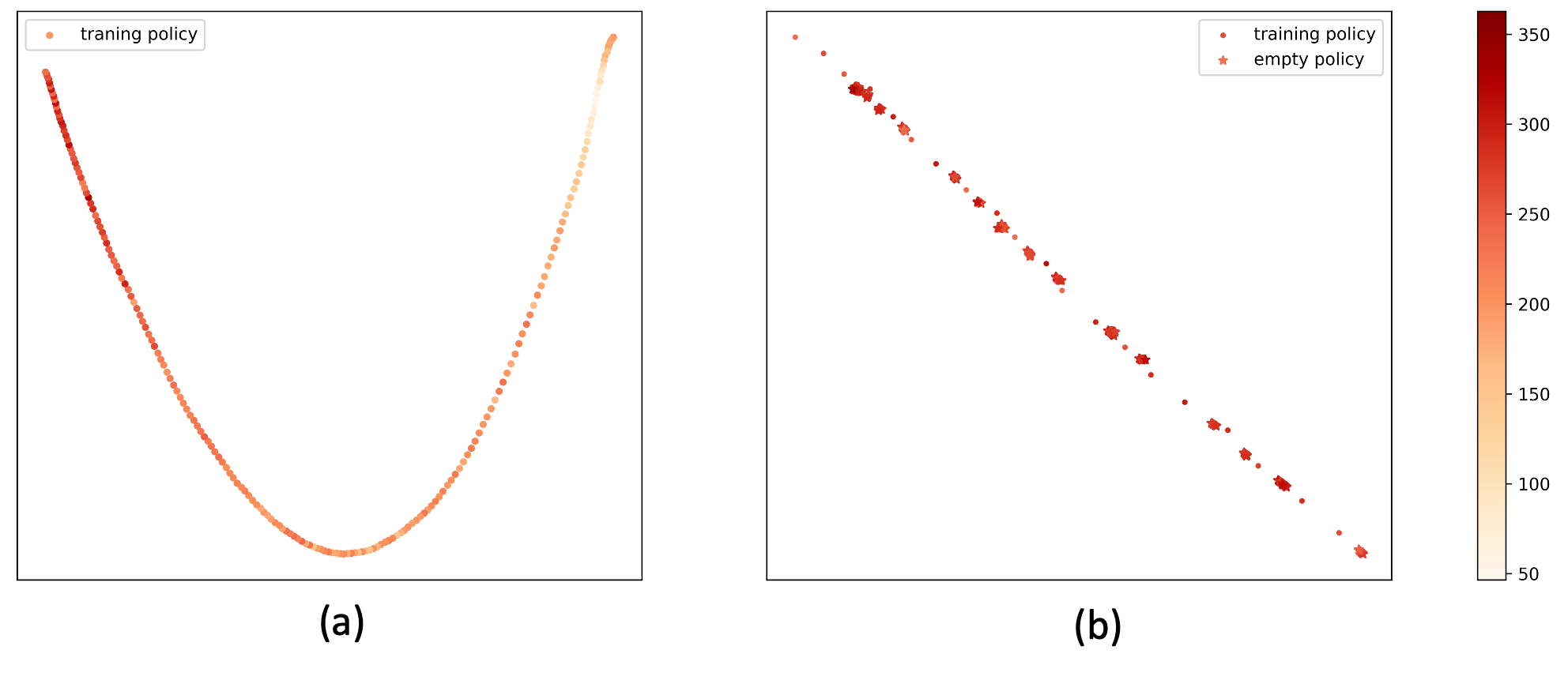}
  	\caption{Policy networks visualized by PCA. The color encodes the policy's return
          averaged over 10 episodes. (a) Policy checkpoints
          from all the training iterations. Each point represents one
          training policy at a checkpoint. 
          (b) Empty space policies and their neighbor training
          policies. Here, points represent training
          policies and stars represent empty ones. This is a local
          space of the last 20 training policies (the top left
          corner in (a)) since the empty-space policies were identified among
          the last 20.}
  \hfill%
  \vspace{-15pt}
  \label{fig:policy_traj}
\end{figure}

\subsection{Systems Dataset}
Fig. \ref{fig:fullviewsystem} shows the full view of GapMiner for the application example described in Fig. \ref{sec:application}, the systems dataset.
\begin{figure*}[tbp]
  \centering
  \includegraphics[width=\linewidth]{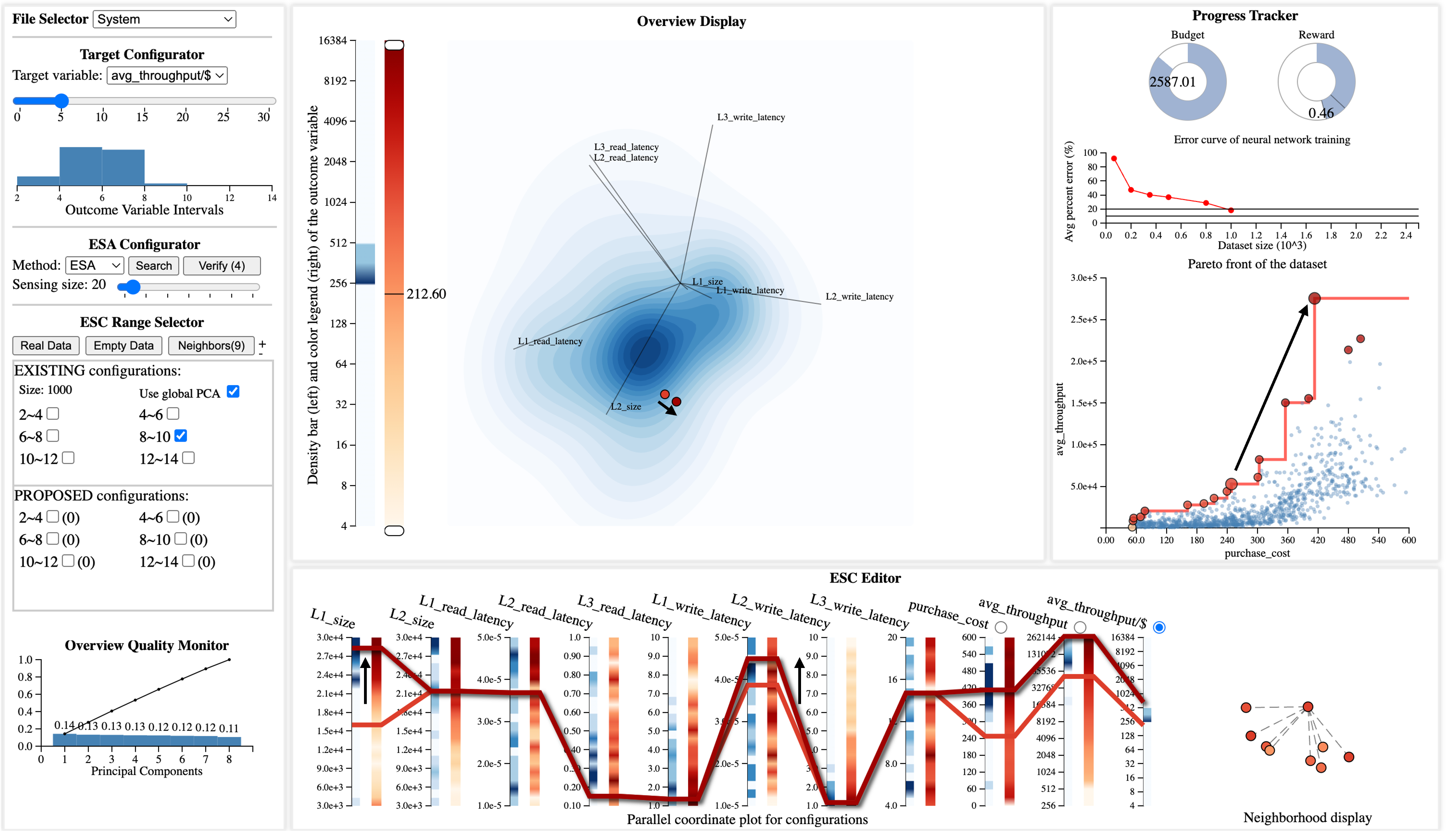}
  	\caption{The full view of GapMiner for the system dataset. The black arrows in the overview display, the PCP, and the Pareto frontier plot show the direction of optimization.}
  \hfill%
  \label{fig:fullviewsystem}
\end{figure*}

\subsection{Wine Investigation}
\label{appendix:wine}
Fig. \ref{fig:fullviewwine} show the full view of GapMiner for the wine dataset. 

\begin{figure*}[tbp]
  \centering
  \includegraphics[width=\linewidth]{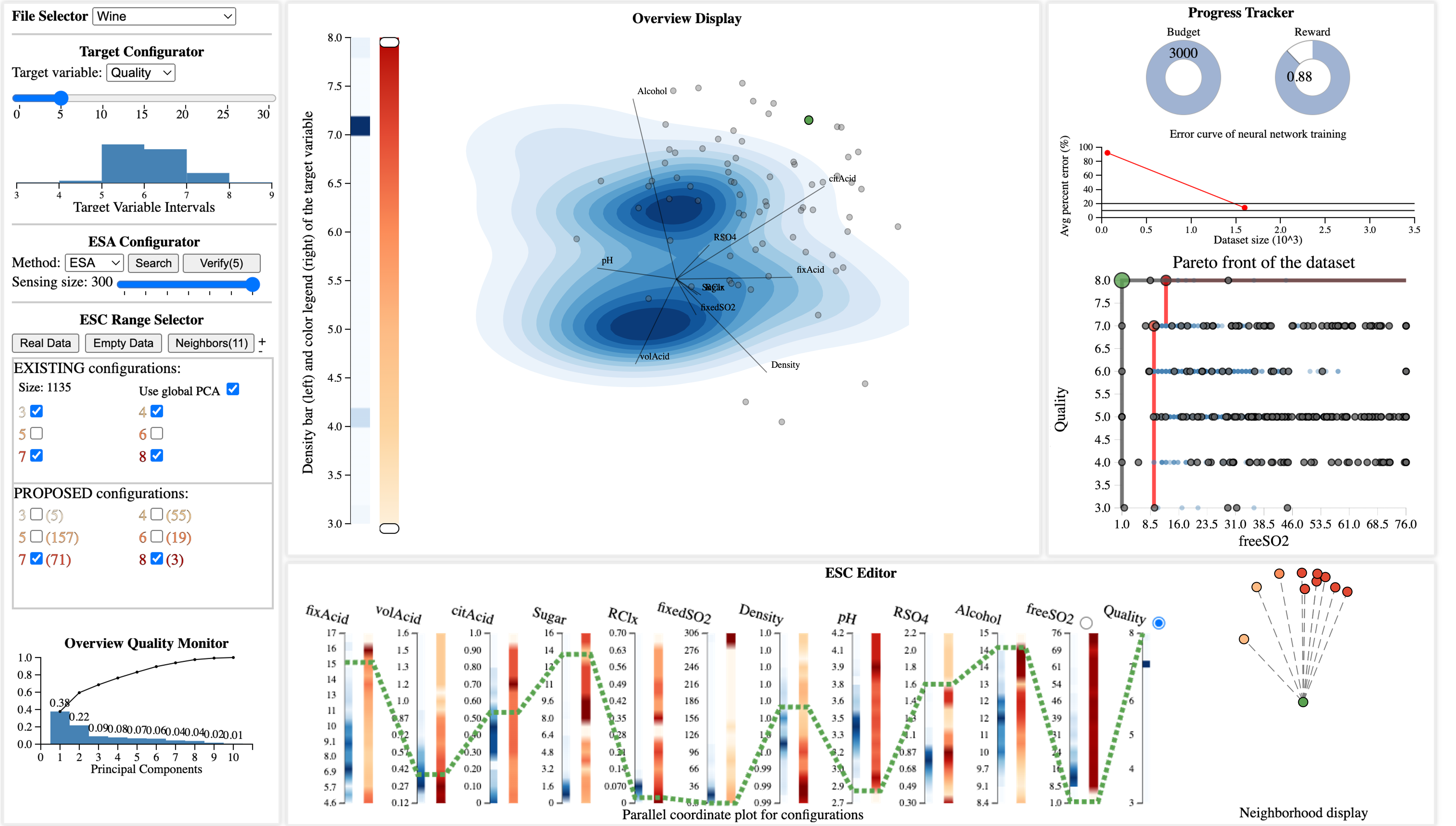}
  	\caption{The full view of GapMiner for the wine dataset study. We highlight the global optimal wine instance in green in the PCA map, the PCP, and the neighbor plot.}
  \hfill%
  \label{fig:fullviewwine}
\end{figure*}

\subsection{Empty Space Images in Adversarial Learning}
\label{appendix:adversarial}

Our first step was to choose relevant digit classes from the control panel and
learn the distribution in latent space by using the PCA map and the
PCP. With this in hand, we explored the latent space by manipulating a line in
the PCP. When the latent point moves in the PCA map, we reconstructed
the image with the decoder and visualized it in the neighbor
plot. Fig. \ref{fig:casemnist}(a) shows an example of the empty-space
exploration. Starting from a digit 7, we can see that all of its neighbor
images are 7 except one 9, and all are close to the empty space image
(dashed line).  Changing the image a bit in the latent space results
in a shallow line at the head, making it a 9, but the CNN fails to
recognize it, still classifying it as a 7. Fig. \ref{fig:casemnist}(b)
shows another adversarial image by ESA, which is a 7 misclassified as
a 5.

Additionally, from Fig. \ref{fig:ES_adv}(b) we can see that tSNE can
correctly embed an adversarial image from
Fig. \ref{fig:casemnist}(a) into cluster 9 (the pink circle in cluster 9),
but it fails for the image from Fig. \ref{fig:casemnist}(b) (the yellow circle
in cluster 5).

\subsection{Empty-Space Search in Reinforcement Learning}
\label{appendix:rl}

We visualize the training policies in the PCA space in
Fig. \ref{fig:policy_traj}(a), and the empty-space policies in the PCA space
in Fig. \ref{fig:policy_traj}(b).

\subsection{Cheetah Direction}

We demonstrate an additional application of GapMiner in a
physical-simulation scenario.

CheetahDir is an environment in MUJOCO that controls a half cheetah robot to move forward. The reward function is composed of two
components: forward reward and control cost. Forward reward is the
velocity in the given direction, the faster the better. Control
cost is the rooted squared sum of the action, indicating the cost to
execute the action, the smaller the better. The total reward is the
forward reward minus the
control cost. The action space has 6 variables: back thigh, back shin,
back foot, front thigh, front shin and front foot. All of them are
numerical and continuous.

The dataset is collected by an expert agent. Given an observation, it
samples actions from its output distribution randomly. In this
scenario, we explored the ability to find better empty space actions
than the expert agent given the observation. The visual analytics can
be found in Fig. \ref{fig:gapminercheetahdir}.

\begin{figure*}[tbp]
  \centering
  \includegraphics[width=\linewidth]{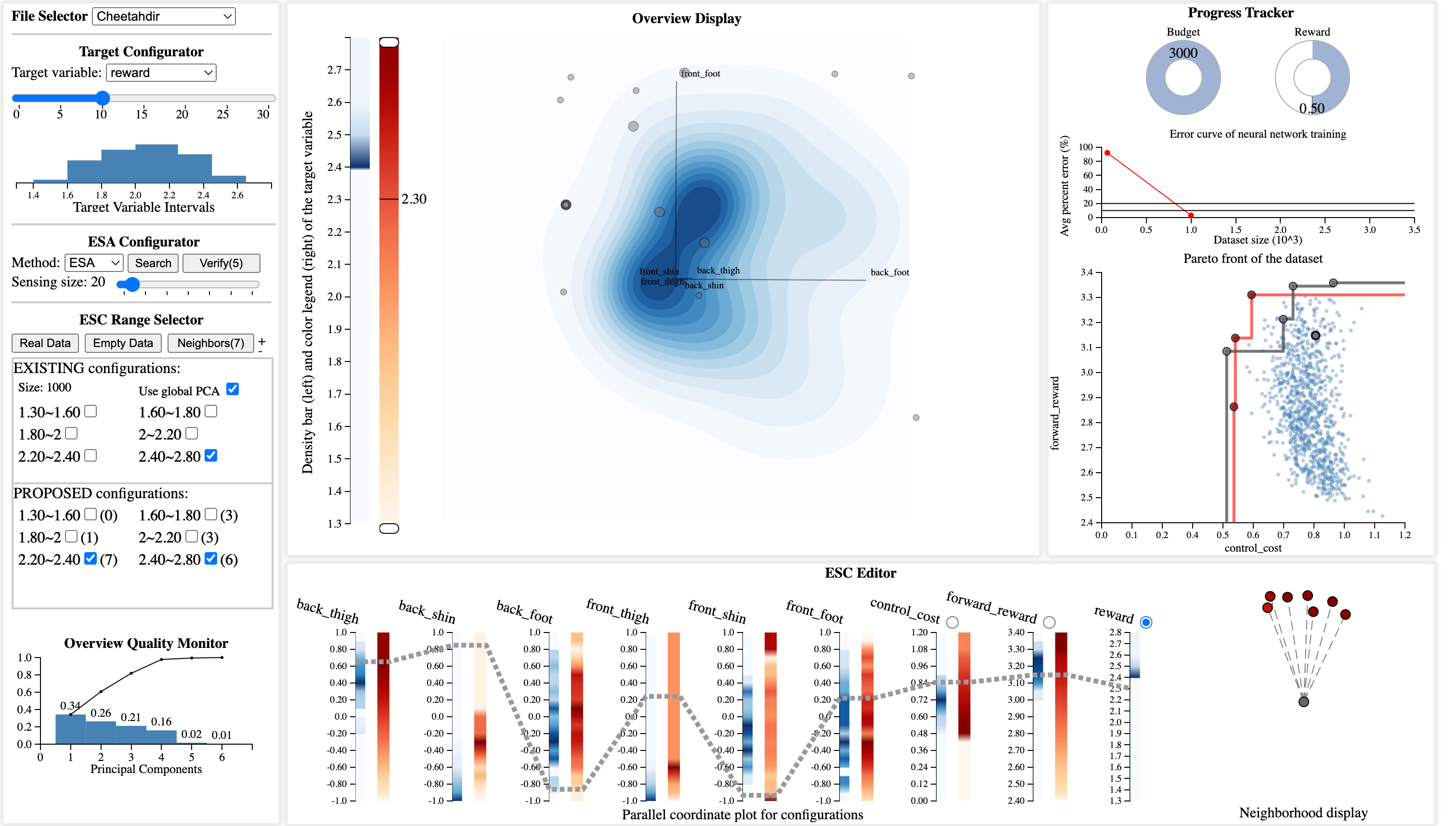}
  	\caption{We trained a neural network to predict forward reward
          and calculate the control cost; then we calculate the reward of an
          action as forward reward minus control cost. We applied the
          dataset to GapMiner, from which we could see a distribution
          shift from low-reward actions to high-reward actions. It
          indicated that front foot and back foot are more important
          to the reward. Then we ran ESA to search empty-space actions
          and plot the predicted forward reward and control cost in the
          Pareto plot. It showed that actions from the expert agent
          always have high cost, but some don't have high reward. ESA
          can find actions that are either cheaper while maintaining
          good reward, or expensive but even higher reward. Notice
          that RL is a sequential decision-making problem. A high
          immediate reward doesn't necessarily imply a long-term
          benefit. Searching
          empty-space actions with higher Q values is another
          interesting application scenario.}
  \hfill%
  \label{fig:gapminercheetahdir}
\end{figure*}
}

\end{document}

%% file: abstract.tex
\begin{abstract}
  %
We present a comprehensive pipeline, augmented by a visual analytics
system named ``GapMiner'', that is aimed at exploring and exploiting untapped
opportunities within the empty areas of high-dimensional datasets.
Our approach begins with an initial dataset and then uses a novel
Empty Space Search Algorithm (ESA) to identify the center points of
these uncharted voids, which are regarded as reservoirs containing potentially
valuable novel configurations.
Initially, this process is guided by
user interactions facilitated by GapMiner.  GapMiner visualizes the
Empty Space Configurations (ESC) identified by the search within the
context of the data, enabling domain experts to explore and adjust
ESCs using a linked parallel-coordinate display.  These interactions
enhance the dataset and contribute to the iterative training of a
connected deep neural network (DNN).  As the DNN trains, it
gradually assumes the task of identifying high-potential ESCs,
diminishing the need for direct user involvement.  Ultimately, once
the DNN achieves adequate accuracy, it autonomously guides the
exploration of optimal configurations by predicting performance and
refining configurations, using a combination of gradient ascent
and improved empty-space searches.  Domain users were actively engaged
throughout the development of our system.  Our findings
demonstrate that our methodology consistently produces substantially
superior novel configurations compared to conventional
randomization-based methods.  We illustrate the effectiveness of our
method through several case studies addressing various objectives,
including parameter optimization, adversarial learning, and
reinforcement learning.
\end{abstract}


%% file: intro.tex
\section{Introduction}
\lettrine[lines=2,lhang=0.,nindent=0em]{T}{HIS}
paper focuses on a methodology for effectively discovering “empty spaces”—regions where data points are absent—in multivariate and  high-dimensional datasets. Identifying and exploring these empty spaces is both a challenge and an opportunity. The challenge is rooted in the \textit{curse of dimensionality}, which is the exponential increase in volume and data sparsity associated with adding dimensions \cite{bellman1966dynamic}. It makes the search for meaningful empty spaces increasingly complex, even for just a moderate number of attributes parameterizing the data.


\sloppy
Overcoming this complexity is not merely academic. It directly
impacts the practicality of discovering and verifying new, unknown, and
yet unimagined
configurations that might reside within these empty
spaces.
Advocating for a configuration-discovery technique that can explore unconventional or even radical changes brought by unknown configurations and parameter settings offers a powerful alternative to conventional parameter tuning and optimization. These tasks are increasingly recognized as high priorities across diverse fields, including aerospace engineering~\cite{brunton2021data}, manufacturing~\cite{cioffi2020artificial}, computer systems~\cite{alipourfard2017cherrypick}, personalized healthcare~\cite{zhang2018learning}, and others, where techniques of this sort often address optimization problems with multiple objectives. 







The emergence of crossover cars and hybrid vehicles illustrates well the high potential of exploring large parameter spaces to discover unique and unexpected configurations.
Crossovers blend features from different vehicle types, offering drivers the versatility of an SUV with the agility of a sedan. Similarly, hybrids integrate gasoline engines with electric motors, improving fuel efficiency and reducing emissions without sacrificing performance. These examples demonstrate how investigating gaps within extensive parameter spaces can challenge conventional design constraints and embrace unconventional configurations. Clearly, innovation of this sort can also occur in lesser contexts.




While ingenious configurations become obvious once discovered, there are
a multitude of them that defy practicality.  Also, more often than not,
high cost and substantial effort are required to obtain or simulate a
hypothesized configuration to verify its merit.  A human expert is often the best judge to decide whether to take up the risk of engaging
in such a testing effort at all.  But even with the human in the
loop, the challenge lies in ideating meritorious configurations in the
presence of this massive realm of possibilities.

To illustrate this challenge, consider a 4-dimensional parameter space with 50 levels for each dimension. This results in $50^4 =$ 6.25 million possible configurations. Suppose we have data on the merit of 10,000 configurations from previous experiments. This means we have information on only 10,000/6,250,000 = 0.16\% of the parameter space. While it is impractical to expect valid data at every location of the parameter space, the challenge lies in identifying \textit{which} configurations are most useful. This uncertainty underscores the importance of intelligent sampling and discovery techniques to uncover valuable and innovative solutions.

While AI holds promise to replace human experts in this search, it requires ample high-quality training data to be effective. Our research supports this role of AI. Inspired by human-in-the-loop (HITL)~\cite{mosqueira2023human} machine learning, our pipeline integrates the training of a deep neural network that evolves alongside the exploration process. This network aids in evaluating identified configurations and improves with accumulated data, enhancing search efficiency for verifying new configurations.



Conceptually, our method looks for gaps in high-D data spaces. While in theory the pairwise distances among data points in high-D space
tend to be normally distributed with small variance, in practice,
however, data configurations aggregate into \textit{hubs}---points
that occur more often in k-neighborhoods of nearby points than
others~\cite{zimek2012survey}.  Likewise, there are also
\emph{anti-hubs}---points that are unusually far away from most other
points~\cite{radovanovic2010hubs}. When these points exist, they are
referred to as outliers, hiding in gaps and sparsly occupied pockets
of the data space.  Essentially, our method looks for
\emph{hypothetical} outliers---thus-far unsampled configurations that
might bear promise or are adversarial.

The major contributions of our paper are hence as follows:
\begin{itemize}
%
\item A scalable, parallelizable \emph{Empty Space Search Algorithm
  (ESA)} that can identify empty spaces in any high-dimensional
  dataset.
\item A visual analytics system, \emph{GapMiner}, by which
  users can further modify the identified Empty Space Configurations
  (ESC).
\item A Human-in-the-Loop (HITL) to AI pipeline that trains an AI
  agent (a DNN) for eventual ESC search autonomy.
\item A dimension-reduction method that allows visualization of
  the neighbor distributions around empty spaces.
\item Several case studies that demonstrate the effectiveness of our
  methodology in diverse application domains and objectives.
\item A user study that evaluates our methodology and implementation.
\end{itemize}


In the following,
Sec. \ref{sec:related_work} presents related work.  Sec. \ref{sec:es_algorithm}
describes our empty space search algorithm. Sec. \ref{sec:ui_design}
introduces our visual analytics system, GapMiner. Sec. \ref{sec:pipeline}
explains our Human-in-the-Loop to AI pipeline.
Sec. \ref{sec:application} showcases our method applied to the computer systems domain and Sec. \ref{sec:userstudy} presents an
associated user study.  Sec. \ref{sec:case} describes three further
case studies. Sec. \ref{sec:discussion} concludes and discusses future work.


%% file: related-work.tex
\section{Related Work}
\label{sec:related_work}

The research literature on the specific topic of empty space visualization is sparse; we only know of two research groups who tackled this. 

\label{sec:esvis}
Strnad \etal~\cite{strnad2013real} investigated the identification of
empty spaces in protein structures. However, their study was limited
to 3D space and did not address the concept of empty spaces in other
fields. Additionally, their algorithm, which relies on Delaunay
Triangulation, faces scalability issues in higher dimensions, as
discussed in Sec. \ref{sec:compugeo}.

Giesen \etal~\cite{giesen2017sclow} introduced the \textit{Sclow plot}
for identifying and visualizing empty spaces. They use flow lines to
depict these empty spaces and employ a scatter-plot matrix to provide
a comprehensive view of the high-D dataset along with the
flow lines. 
A downside of this approach, however, is that the flow lines become cluttered and difficult to track and explain at large scales
and the scatter-plot matrix view suffers from quadratic growth with increasing data
dimensionality, which further complicates the visualization. Also, Sclow plots focus on detecting data distribution features, while our work concentrates on applying empty space points for optimal configuration search and multi-objective optimization. 



We separate the remaining related research into three areas: (1) high-D visualization, (2) identification of empty space via computational geometry, and (3) optimal configuration search.

\subsection{High Dimensional Space Visualization}
\label{sec:dr}



Understanding empty spaces within high-D environments relies on
effective visualization. While various dimension-reduction methods
have been devised to project high-D datasets onto a 2D screen, it is
important to note that dimension reduction capitalizes on the
existence of empty spaces, deflating them to pack the existing points
as efficiently as possible in the 2D display. Therefore it is best to
conduct empty-space analysis in the native high-D space, with only
redundant dimensions removed.

Prominent linear techniques include Principal Component Analysis (PCA)
\cite{hotelling1933analysis}, Linear Discriminant Analysis (LDA)
\cite{cohen2013applied}, and classical Multidimensional Scaling (MDS)
\cite{mead1992review}, all of which project data to a
lower-dimensional space while preserving global structure and
minimizing distortion. However, these methods often struggle to
capture the complexities of data manifolds.
On the other hand, manifold-learning techniques such as Locally Linear
Embedding (LLE) \cite{roweis2000nonlinear}, Uniform Manifold
Approximation and Projection (UMAP) \cite{mcinnes2018umap}, and
t-distributed Stochastic Neighbor Embedding (tSNE)
\cite{van2008visualizing} excel at revealing nonlinear relationships
and intrinsic data geometries but the embedding process loses the
context of the attributes. 
The Data Context Map (DCM) \cite{cheng2015data} offers a unified view for visualizing both variables and data items. Zhang \etal \cite{zhang2022graphical} enhanced DCM with graphical contours to identify exemplars, while Bian \etal \cite{bian2020implicit} improved dimension-reduction methods by using glyphs to denote intrinsic dimensions in low-D representations.



Parallel-Coordinate Plots (PCPs) \cite{inselberg1985plane}  directly explore the original data space, thus avoiding potential information loss and distortion. PCPs arrange dimensions linearly, aiding in uncovering data relationships and patterns. 
But PCPs are not without challenges, which has inspired efforts to
reduce visual clutter~\cite{ellis2007taxonomy},
enhance subset tracing and correlation through bundle
representation~\cite{palmas2014edge}, employ graphical
abstraction~\cite{mcdonnell2008illustrative}, and introduce a
many-to-many axis format to reveal deeper variable
relationships~\cite{lind2009many}.
We combine PCPs with PCA to visualize the essential attributes of
high-D datasets.  This integrated approach synergizes the strengths of
both techniques, providing a comprehensive, user-centric
exploration of high-D spaces in a cohesive and interactive manner.

Subspace analysis is yet another technique for dimension
reduction~\cite{kriegel2012subspace}.  It decomposes the high-D data
space into a set of lower-D subspaces.  This can help reveal data
patterns obscured by irrelevant (\eg noisy) data dimensions.  A
challenge here is the combinatorial explosion in the set of possible
subspaces, and this has been the subject of extensive
research~\cite{ mahmood2023interactive, 
  wang2017subspace, wang2019high, tatu2012clustnails}.  Our method could readily incorporate these techniques, and we plan to study this in future work.


\subsection{Empty-Space Identification in Computational Geometry}
\label{sec:compugeo}
One approach to identify empty spaces is from a geometrical
perspective.  Computational-geometry techniques offer a complete
partition of the data space based on the dataset, and this can be used
to  locate empty regions accurately among points.  These techniques use
Delaunay Triangulation, Voronoi Diagrams, and Convex Hulls, all of which are
interconnected: the $d$-dimensional Delaunay Triangulation corresponds
to the $(d+1)$-dimensional convex hull;  Delaunay Triangulation and
Voronoi Diagrams are in duality,
\ie the circumscribed circle centers of Delaunay triangles serve as
vertices of Voronoi Diagrams~\cite{de2000computational}.

We initially studied these techniques with a specific focus on
Delaunay Triangulation.  Each circumscribed circle of a Delaunay
triangle describes an empty space, facilitating a comprehensive
exploration.  However, the downside is its exponential time and space
complexity, both $O(n^{\ceil{{d/2}}})$~\cite{seidel1995upper}. 
Efforts to mitigate time
complexity constraints have explored parallel-computing acceleration,
which has shown promise in 2D and 3D spaces~\cite{blelloch1999design,
  nguyen2020delaunay}.  
  However, extending these
methods to higher dimensions is non-trivial and this limits their
utility in analyzing empty spaces within multivariate datasets. 
App. \ref{appendix:ESA-Delaunay}
presents empirical studies we conducted that reveals these shortcomings.

Other solutions include approximation methods.  Peled
\etal~\cite{har2001replacement} proposed a Voronoi Diagram replacement
with linear space complexity.  Balestriero
\etal~\cite{balestriero2022deephull} introduced Deep Hull, which
employs a deep neural network to determine points within or outside
the convex hull.  Graham \etal~\cite{graham2017approximate} proposed a
method for higher-dimensional convex hull identification, but its
unordered points impede the transferability of Delaunay triangulation.
Although these methods reduce time or space complexity, they either
approximate specific attributes or lack a meaningful order, limiting
their applicability.

\subsection{Optimal Configuration Search}
A defining goal of ours is optimal configuration search, which
involves systematically navigating through various configurations,
arrangements, or settings within a system, application, or model to
identify those that deliver the best performance based on specific
criteria or objectives.


Optimizing an algorithm or machine-learning model's hyperparameters is often complex. Grid search, which equally divides the range of each variable and examines every point within the high-dimensional data space, is intuitive but inefficient. Alternative strategies like Iterated Local Search (ILS) \cite{lourencco2003iterated} and its enhancements, such as ParamILS and Focused ILS \cite{hutter2007automatic}, use hill climbing to iteratively adjust and improve the current solution. 
Most recent methods rely on population based optimization to explore the hyperparameters space efficiently~\cite{Parker-Holder2020,Japa2023}.
These methods require many evaluations to effectively explore the solution space. Our approach, in contrast, focuses on directly identifying and probing candidate solutions that are most distant from existing ones, aiming to find new data instances outside the current range.

Deep neural networks (DNNs) have also been employed for optimal configuration search, serving as function approximators to expedite configuration optimization \cite{ villarrubia2018artificial, chen2022neural}. 
While powerful, DNN-based approaches require large datasets for training to ensure reliable results, which can be challenging when working with limited data.

